\documentclass{article}

\usepackage{multicol}
\usepackage[bookmarks=true]{hyperref}

\usepackage{amsmath}
\usepackage{amssymb}
\usepackage{mathtools}
\usepackage{amsthm}
\usepackage{amsfonts}
\usepackage{algorithm}
\usepackage[noend]{algpseudocode}

\usepackage[dvipsnames]{xcolor}
\usepackage{color}

\usepackage{bm}
\usepackage{bbm}
\usepackage{diagbox}
\usepackage{wrapfig}
\usepackage{soul}
\usepackage{comment}
\usepackage{subcaption}
\usepackage{array,booktabs}
\usepackage{multirow}
\usepackage{adjustbox}
\usepackage{graphicx}
\usepackage{tabularx}
\usepackage{wrapfig}
\usepackage[title]{appendix}
\usepackage{listings}

\usepackage{cleveref}
\crefname{section}{Sec.}{Secs.}
\crefname{subsection}{Sec.}{Secs.}
\crefname{subsubsection}{Sec.}{Secs.}
\crefname{equation}{Eq.}{Eqs.}
\crefname{figure}{Fig.}{Figs.}
\crefname{algorithm}{Alg.}{Algs.}

\usepackage{enumitem}
\newcommand{\assumpitem}[2]{(#1#2)}

\usepackage[preprint]{conference}

\title{\mbox{\includegraphics[height=1.8ex]{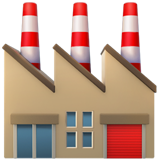} Fabrica: Dual-Arm Assembly of General Multi-Part} Objects via Integrated Planning and Learning}

\addtocounter{footnote}{1} 
\author{
Yunsheng Tian$^{1,}$\thanks{Correspondence to: \texttt{yunsheng@mit.edu}, \texttt{hui.xylo.li@autodesk.com},
and \texttt{wojciech@mit.edu}}, 
Joshua Jacob$^{1}$,
Yijiang Huang$^{2}$,
Jialiang Zhao$^{1}$,
\\ \textbf{
Edward Gu$^{1}$,
Pingchuan Ma$^{1}$,
Annan Zhang$^{1}$,
Farhad Javid$^{3}$,
Branden Romero$^{1}$,
}
\\ \textbf{
Sachin Chitta$^{3}$,
Shinjiro Sueda$^{4}$,
Hui Li$^{3,\dagger}$,
Wojciech Matusik$^{1,\dagger}$}\\
$^1$MIT CSAIL, $^2$ETH Zurich, $^3$Autodesk Research, $^4$Texas A\&M University
}

\begin{document}

\definecolor{junglegreen}{rgb}{0.16, 0.67, 0.53}
\newcommand{\citeme}[1]{{\bfseries \scriptsize  \color{junglegreen}\framebox{CITE: {\color{black} #1}}}}
\newcommand{\todo}[1]{{\bfseries \scriptsize  \color{red}\framebox{ToDo: #1}}}
\newcommand{\addref}[0]{{\bfseries \scriptsize  \color{junglegreen}\framebox{REF}}}
\newcommand{\drawme}[1]{\centerline{\scriptsize\centering \color{cyan}\framebox{ \textsc{Draw me:} {\color{black} #1}}}}

\newcommand{\yunsheng}[1]{{\textcolor[rgb]{0.7,0.2,0.2}{Yunsheng: #1}}}
\newcommand{\YH}[1]{{\textcolor[rgb]{0.2,0.2,0.7}{YH: #1}}}

\newcommand{\hui}[1]{{\textcolor[rgb]{0.2,0.8,0.2}{Hui: #1}}}

\setlength{\abovedisplayskip}{6pt}
\setlength{\belowdisplayskip}{2pt}

\setcounter{figure}{1}
\makeatletter
\let\@oldmaketitle\@maketitle%
\renewcommand{\@maketitle}{
   \@oldmaketitle%
\begin{center}
    \vspace{-7mm}
    \centering
    \includegraphics[width=\textwidth]{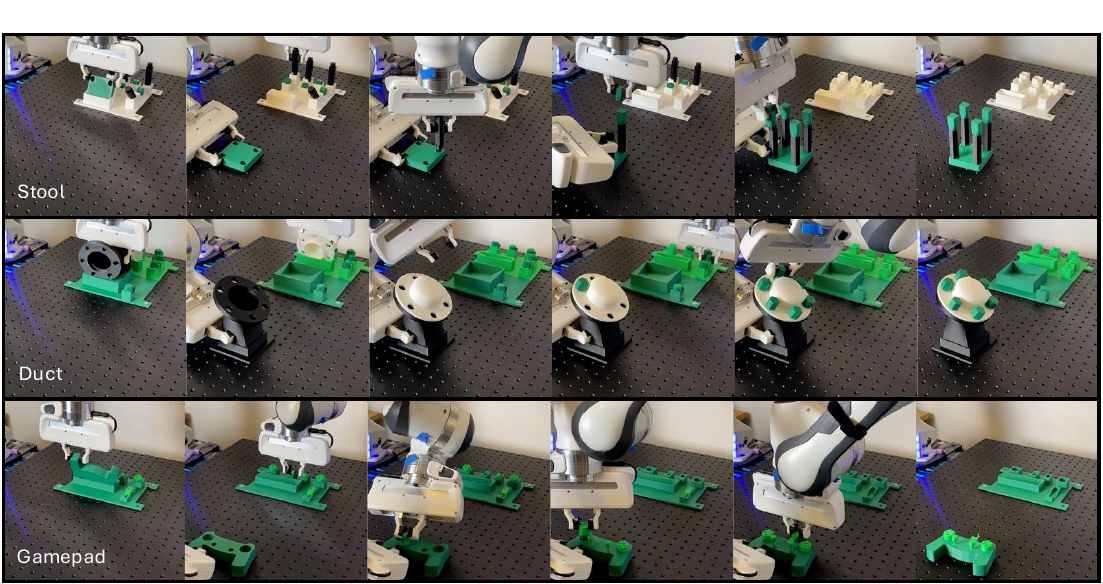}
\end{center}
\vspace{-3mm}
    Fig.~\thefigure:\label{fig:teaser} {Our proposed dual-arm robotic system demonstrates adaptive manipulation and assembly capabilities for diverse multi-part objects. The system combines offline task-oriented planning and optimization to address sequencing, grasping, and motion planning for long-horizon assembly tasks. For robust online control, it utilizes guidance from the offline plan to learn assembly skills that generalize effectively across diverse object geometries, assembly paths, and grasp poses.
    }
}
\makeatother 

\maketitle


\begin{abstract}
Multi-part assembly poses significant challenges for robots to execute long-horizon, contact-rich manipulation with generalization across complex geometries. We present Fabrica, a dual-arm robotic system capable of end-to-end planning and control for autonomous assembly of general multi-part objects. For planning over long horizons, we develop hierarchies of precedence, sequence, grasp, and motion planning with automated fixture generation, enabling general multi-step assembly on any dual-arm robots. The planner is made efficient through a parallelizable design and is optimized for downstream control stability. For contact-rich assembly steps, we propose a lightweight reinforcement learning framework that trains generalist policies across object geometries, assembly directions, and grasp poses, guided by equivariance and residual actions obtained from the plan. These policies transfer zero-shot to the real world and achieve 80\% successful steps. For systematic evaluation, we propose a benchmark suite of multi-part assemblies resembling industrial and daily objects across diverse categories and geometries. By integrating efficient global planning and robust local control, we showcase the first system to achieve complete and generalizable real-world multi-part assembly without domain knowledge or human demonstrations. Project website: \href{http://fabrica.csail.mit.edu/}{fabrica.csail.mit.edu}
\end{abstract}

\keywords{Assembly, Planning, Reinforcement Learning, Benchmark} 


\section{Introduction}
Multi-part assemblies are prevalent in home and industrial settings.
Robotic assembly of multi-part objects presents a longstanding challenge: long-term planning to map CAD models to robot programs and robust control skills to achieve high precision and adaptivity during contact-rich interactions.
However, most assembly robots today are programmed manually with specially designed infrastructures, and the program is executed repetitively using a stiff controller.
As a result, they take substantial time to adapt to new production demands and are highly sensitive to uncertainties.

Despite recent progress in sim-to-real transfer of contact-rich part insertion skills~\cite{xu2022efficient, zhang2024bridge, tang2024automate}, current robotic systems are still not capable of assembling general multi-part objects.
Prior research has primarily focused on two-part, top-down insertion using a single robot arm, but multi-part assembly requires diverse insertion and grasping poses and a bi-manual operation that frequently changes which part to hold to counter-balance the insertion force from the other hand.
This presents new challenges to planning and control.
First, jointly finding an assembly-hold sequence, physically stable grasps, and collision-free robot motion presents a hybrid (discrete-continuous) optimization problem in a large search space.
Second, control policies for part insertion must be robust to misalignment and uncertainty, while being able to generalize across a wide range of part geometries.

We tackle these challenges by building a general planning and control system for flexible, dual-arm assembly of multi-part objects, with zero-shot sim-to-real transfer. 
Our contributions include:

{\bf Algorithms}: We propose a hierarchical dual-arm planner to plan and optimize the assembly-hold sequence, grasps, and robot motion.
For contact-rich steps, we learn generalist reinforcement learning (RL) policies utilizing equivariant representations guided by planned motion to achieve robustness.

{\bf Systems}: We build a real-world system that can map a CAD assembly model to robot execution that alternates between tracking planned motions and reactive control policies. 
To our knowledge, this is the first system that autonomously achieves all phases of a multi-part assembly problem: from automatic pickup fixture design, to sequence, grasps, and motion planning, to insertion. Our system is tested on commonly used robotics hardware and can be generalized to different dual-arm robots.

{\bf Benchmarks}: We design a benchmark suite of 7 multi-part assemblies ranging from 5 to 9 parts, and our system can assemble them robustly in both simulation and real-world system.

\section{Related Work}

Prior work on multi-part assembly is heavily focused on planning assembly sequences and paths, including geometric reasoning~\cite{halperin2000general}, sampling-based motion planners~\cite{sundaram2001disassembly, le2009path, zhang2020c}, and RL for combinatorial sequence search~\cite{funk2022learn2assemble, ghasemipour2022blocks}. Recently, physics-based motion planning~\cite{tian2022assemble} has shown success in assembling many complex parts with tight clearances. In addition, realistic kinematic and dynamic constraints have been considered in sequence planning for real-world robot setups~\cite{tian2024asap, rodriguez2019iteratively, zhu2024multi}. However, planning alone struggles with execution uncertainties, and stability- or efficiency-optimal plans remain underexplored.
While robust and efficient robotic systems have been built for tasks like assembling IKEA chairs~\cite{suarez2018can}, LEGO blocks~\cite{nagele2020legobot}, and structural elements~\cite{huang2021robotic, huang2021new, wang2023temporal, liang2017ras, dorfler2016mobile, apolinarska2021robotic, kramberger2022robotic}, these are domain-specific and lack generalizability. In contrast, our planner generalizes across diverse multi-part assemblies, employs hierarchical structure and parallelization for efficiency, and explicitly optimizes stability to enhance downstream control robustness.

Even with given assembly plans, executing contact-rich assembly remains challenging due to tight clearances, system uncertainties, and the need for generalization. RL has shown promise in addressing these issues, combining motion planning with policy learning from CAD models or supervised trajectories~\cite{thomas2018learning, fan2019learning}, and leveraging accurate simulations for motion generation and policy training~\cite{tian2024asap, narang2022factory}. Sim-to-real transfer~\cite{tang2023industreal, tang2024automate, zhang2023efficient, zhang2024bridge} and real-world RL~\cite{luo2024serl} have enabled high-precision insertion, while some efforts~\cite{noseworthy2024forge} explore multi-step tasks. However, they all primarily work under simplified settings like top-down insertions and fixed grasps, which are insufficient for multi-part assembly where side-way insertions or tilted grasps are necessary. Imitation learning approaches~\cite{ankile2024juicer, ankile2024imitationrefinementresidual} support complex multi-step skill learning but lack robustness and generality. While spatial-equivariant techniques have improved generalization in other domains~\cite{huang2024leveraging, simeonov2022neural, simeonov2023se, yang2023equivact, wang2024equivariant}, they remain underexplored for assembly. Notably, \citet{seo2023contact} learn an SE(3)-equivariant gain scheduling policy, but without varying grasps or geometries. Existing benchmarks focus on narrow tasks~\cite{lian2021benchmarking, fan2018surreal, lee2021ikea, heo2023furniturebench, luo2024fmb}, limiting the evaluation of generalization. In contrast, we demonstrate that combining planning with equivariant generalist policies, for the first time, enables multi-part assembly over diverse geometries, paths, and grasps without any human demonstration.

\section{Planning Multi-Step Dual-Arm Assembly}
\label{sec:plan}

\begin{figure*}[ht]
    \centering
    \includegraphics[width=\textwidth]{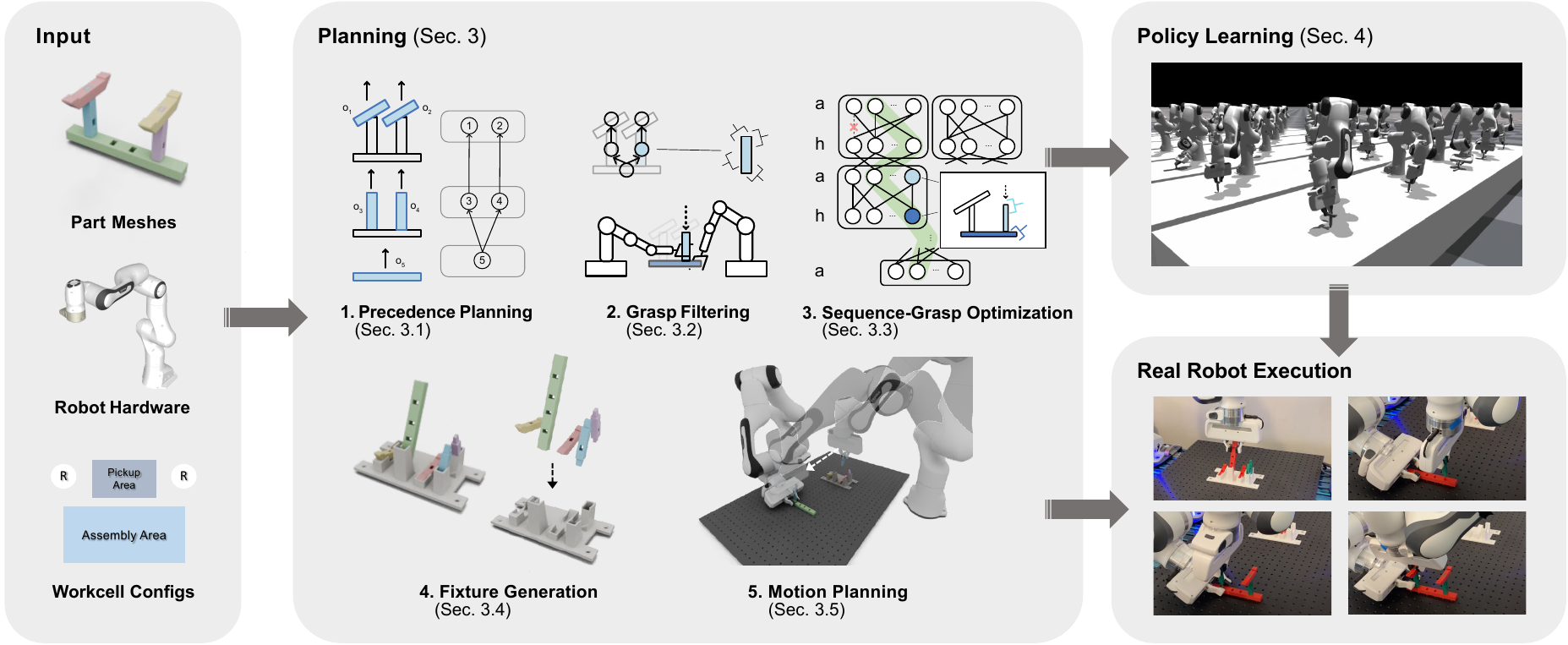}
    \caption{System overview. Fabrica takes part meshes and hardware configurations as inputs. 
    It plans sequences, grasps, fixture designs, and motions through a multi-stage planner, and learns RL policies for all insertion steps, which are deployed together on real robots to complete the assembly.
    }
    \label{fig:overview}
\end{figure*}

Given a $n$-part assembly with parts indexed by $o \in O$, we compute a plan to manipulate all parts from the initial poses $p_O^0$ to the goal poses $p_O^G \in SE(3)$ under all physical constraints.
We focus on sequential, collaborative manipulation that alternates between robot $R_a$ assembling a part and another robot $R_h$ holding a part to stabilize the sub-assembly.
Then, we train control policies for precise contact-rich assembly steps. Finally, our system execution alternates between open-loop planned motions and closed-loop reactive policies.
\Cref{fig:overview} provides an overview of the system.

We formulate planning as optimizing assembly-hold sequences \(\phi\), grasps \(\sigma\), and robot motions \(\pi\):
\begin{equation}
\label{eq:optim}
\min_{\phi, \sigma, \pi} E\big( \Phi_{i=1}^{n}\vec{f}(\phi_{1:i}, \sigma_{i-1:i}, \pi_{i})\big) \quad
\text{s.t.} \quad C_{\text{prec}}(\phi) \leq 0, \hspace{0.2em}
C_{\text{kin}}(\phi, \sigma, \pi) = 0, \hspace{0.2em}
C_{\text{col}}(\phi, \sigma, \pi) \leq 0
\end{equation}
The sequence $\phi = [o_{a,1}, o_{h,2}, o_{a,2}, \cdots, o_{h,n}, o_{a,n}]$ is an ordering of parts to be held ($h$) and assembled ($a$), with $\sigma = [g_{a,1}, g_{h,2}, g_{a,2}, \cdots, g_{h,n}, g_{a,n}]$ including grasp $g \in SE(3)$ for each step.
The robot motion $\pi$ is divided based on the mode families \cite{alami1994two,simeon2004manipulation} and skills: 
$\pi = [\underbrace{\tau_{a,1}^f, \tau_{a,1}^g, \tau_{a,1}^a}_{\pi[a,1]}, 
   \underbrace{\tau_{h,2}^f}_{\pi[h,2]}, 
   \underbrace{\tau_{a,2}^f, \tau_{a,2}^g, \tau_{a,2}^a}_{\pi[a,2]}, \cdots]$
where each assembly task $\pi[a,i]$ for $R_a$ contains (1) a transit motion $\tau_{a,i}^f$ with its hand free, (2) a transfer motion $\tau^g_{a,i}$ grasping an object, and (3) an assembly motion $\tau_{a,i}^a$ for part insertion.
A hold task $\pi[h,i]$ only involves a hold transfer motion $\tau_{h,i}^f$.
The cost function composes step-wise objective vectors \(\vec{f}\) that evaluate the quality of each step, \(\Phi: \mathbb{R}^{|\vec{f}| \times n} \rightarrow \mathbb{R}^{|\vec{f}|}\) aggregates objectives across steps (e.g., sum or max), and \(E\) maps the result to a scalar (e.g., weighted sum). $C_{\text{prec}}, C_{\text{kin}}, C_{\text{col}}$ represent part precedence, kinematic, and collision constraints respectively.
Please refer to App.~\ref{app:plan_formulation} for detailed constraint formulation.

Solving for feasible or even optimal solutions in a joint manner is intractable. We present a hierarchical approach to decompose it into simpler subproblems for efficient computation with optimality guarantees under assumptions~\ref{assump:insert-only}-\ref{assump:finite_grasp}. The complete pesudocode can be found in App.~\ref{app:algo}.

\subsection{Part Precedence Planning}
\label{sec:precedence}

\begin{wrapfigure}{r}{0.4\textwidth}
    \vspace{-12mm}
    \centering
    \includegraphics[width=\linewidth]{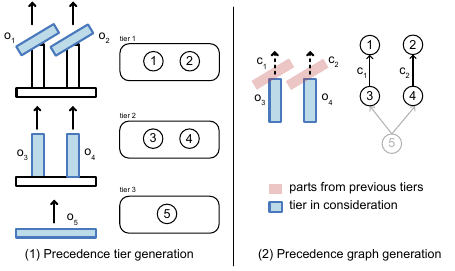}
    \label{fig:precedence_graph_gen}
    \vspace{-8mm}
\end{wrapfigure}

To evaluate constraint $C_\text{prec}$, we propose an algorithm to determine the complete precedence relationships for assembling all parts $O$. First, we define precedence tier as a group of parts that can be removed independently of one another. Tiers are ordered so that parts in earlier tiers must be disassembled before those in later ones. To iteratively construct all tiers, we use a physics-based motion planner~\cite{tian2022assemble} to find all parts that can be disassembled without interfering with the rest, which are grouped into the current tier. We then remove these parts and repeat the process on the remaining assembly until each part is assigned to a tier. 
Next, we build a precedence graph $G_\text{prec}$ that encodes the minimal set of ordering constraints that any collision-free assembly sequence must follow.
Each node in $G_\text{prec}$ is a part, and a directed edge $o_i \rightarrow o_j$ means that $o_i$ must be assembled before $o_j$ because $o_j$ blocks the (dis)assembly path $\tau_{o_i}$ planned during tier generation. For each part $o$, we define its precedence set $O_\text{prec}[o] = \{o' |(o'\rightarrow o) \in G_\text{prec}\}$ as all parts that must be assembled before it.

\subsection{Dual-Arm Grasp Filtering}
\label{sec:grasp}

\begin{wrapfigure}{r}{0.4\textwidth}
    \vspace{-14mm}
    \centering
    \includegraphics[width=\linewidth]{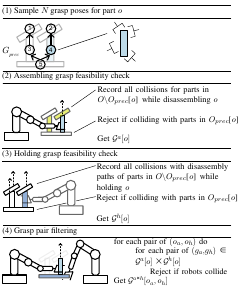}
    \label{fig:grasp_precomp}
    \vspace{-8mm}
\end{wrapfigure}

We aim to identify valid grasp pairs $\mathcal{G}^{a \times h}[o_a, o_h]$ for each assembly-hold part pair $(o_a, o_h)$ which support insertion and holding without colliding with preceding parts of $o_a$ and $o_h$.
Since searching the full 6-DoF space is infeasible, we assume feasible grasps exist in a dense, finite set of grasps, see \ref{assump:finite_grasp}. Because each grasp must be checked for collisions with the current subassembly $\psi_{1:i}$, doing this online results in repeated and expensive checks against many part combinations. To speed up, we precompute valid grasps offline by sampling grasp candidates and performing parallelized inverse kinematics (IK) and collision checks for both arms. 
In practice, we simulate the robot following the part's disassembly path $\tau_{o}$ and reject it if the motion collides with the precedence set $O_\text{prec}[o]$. All collisions and motions are recorded for reuse in later stages. Similarly, we check if a grasp can securely hold a part while disassembling other parts without collision. 
Finally, for each part pair $(o_a, o_h)$, we filter feasible grasps $(g_a, g_h)$ by checking interarm collisions under computed IK.

\subsection{Dual-Arm Sequence-Grasp Optimization}
\label{sec:seq_opt}

\begin{wrapfigure}{r}{0.4\textwidth}
    \vspace{-13mm}
    \centering
    \includegraphics[width=\linewidth]{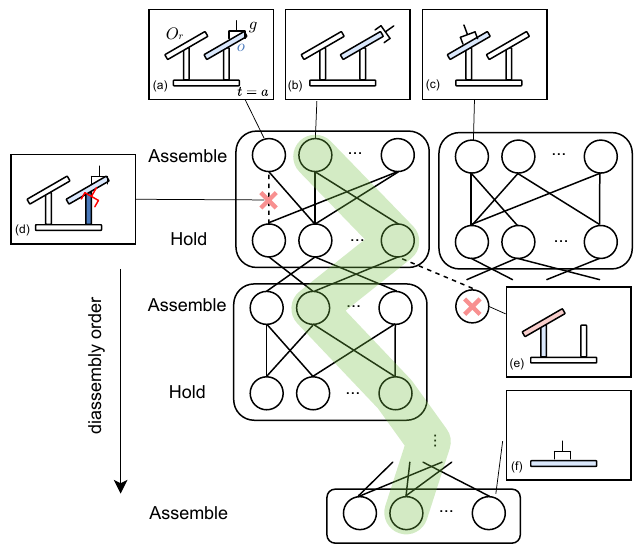}
    \label{fig:grasp_tree}
    \vspace{-8mm}
\end{wrapfigure}

With all valid grasps computed, we now solve for the optimal sequence $\phi^*$ and grasps
$\sigma^*$ in Eq.~\ref{eq:optim}.
We formulate this as a state-space search problem and construct a directed state tree $T_G$, where each node represents a partial assembly state $s = (t, O_r,o,g)$ consisting of robot task $t$ (assemble or hold), assembled parts $O_r$, part being grasped $o$, and grasp pose $g$.
Starting from root nodes (complete assembly), we recursively expand the tree by alternating between assembling and holding, pruning states that violate constraints. Valid transitions must also respect precomputed grasp feasibility between successive steps. All collision and motion feasibility checks are reused from the earlier filtering stage.
Each transition is scored by a grasp stability vector $\vec{f}$, capturing objectives such as supportiveness of the held part, frequency of grasp switches, torque stability, and contact area, which are designed to be lightweight yet effective for downstream control. We apply dynamic programming (DP) to propagate the best cumulative scores through the tree and identify the optimal solution. 
Please find more details in App.~\ref{app:seq_opt}.

\subsection{Grasp-Aware Pickup Fixture Generation}
\label{sec:fixture}

\begin{figure*}[ht!]
    \centering
    \includegraphics[width=\textwidth]{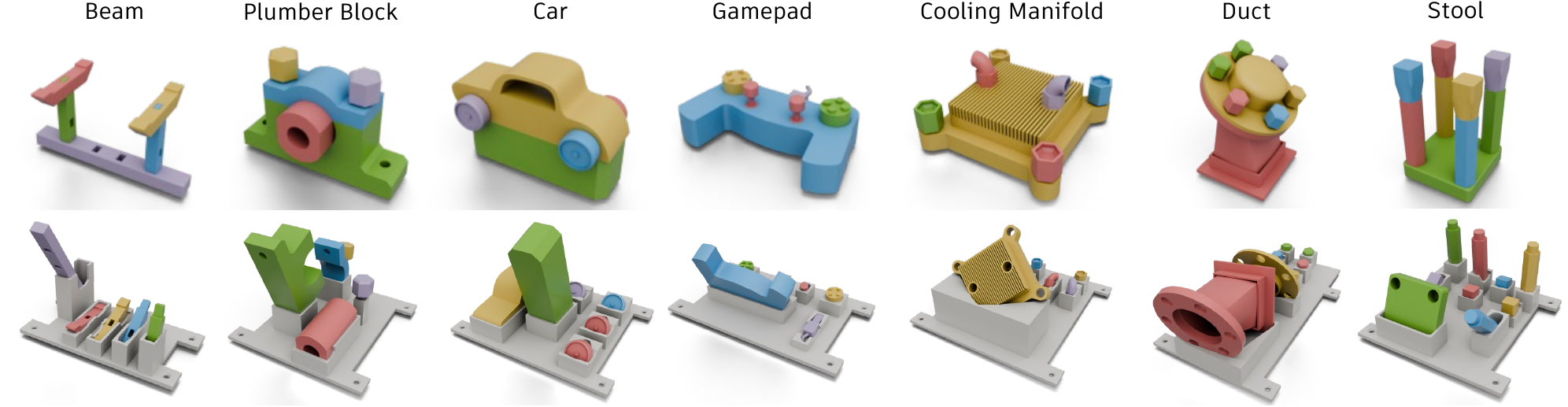}
    \caption{Top: benchmark assemblies. Bottom: the auto-generated pickup fixtures in \cref{sec:fixture}.}
    \label{fig:benchmark}
    \vspace{-3mm}
\end{figure*}

For precise pickup, we develop a software-hardware co-design approach to automatically generate a fixture that stabilizes and orients each part for top-down pickup, based on planned grasps in Sec.~\ref{sec:seq_opt}. This removes the need for reorientation or regrasping between pickup and assembly, allowing the system to focus on the core assembly challenges.
We first determine each part's pickup pose in the world frame, with its orientation defined by the rotation from the assembly grasp to a top-down grasp. We then compute pickup positions by packing parts on the XY plane to avoid collisions between parts and the gripper. To reduce material usage and workspace area, we model this as a bin-packing problem and solve it with an iterative algorithm that alternates between packing, collision checking, and resolution.
Finally, we generate the fixture by creating mold cavities based on the part geometries and poses, ensuring stable placement. See details in App.~\ref{app:fixture} and examples in Fig.~\ref{fig:benchmark}.

\subsection{Motion Planning for Transit and Transfer}
\label{sec:motion_plan}

Finally, RRT-Connect~\cite{kuffner2000rrt} plans for all remaining transit and transfer motions, i.e., $\tau^f_{a,i}, \tau^g_{a,i}, \tau^f_{h,i}$, which can be parallelized since all start and goal states of motions are provided from earlier stages.

\section{Learning General Single-Step Assembly Policy}
\label{sec:learning}

Once the full assembly plan is computed, the next challenge is to track it reliably in the real world. We use a hybrid controller that alternates between tracking the pre-planned transit and transfer motions, and an RL-based reactive controller for contact-rich assembly steps. The controller must generalize across variations in object geometry, grasp poses, and assembly directions. To this end, we design a lightweight yet highly effective RL framework for training a generalist assembly policy.
Given a pre-planned insertion path $\tau_o$, our goal is to guide the part from its noisy initial pose $\hat\tau_o[0]$ to the goal pose $\hat\tau_o[1]$, accounting for uncertainty in grasp and mating part geometry. We frame this as a Markov decision process (MDP) and learn a policy $\pi: \mathcal{O} \to \mathbb{P}(\mathcal{A})$ that maximizes expected rewards over time $\mathbb{E}_{\pi} [\sum_{t=0}^{T-1} \gamma^t r(s_t)]$. We use proximal policy optimization (PPO)~\cite{schulman2017proximal} to train a stochastic policy in simulation, which is then transferred to the real world without additional fine-tuning.

\subsection{Path-Centric Coordinate Transformation} 

Humans naturally reuse the same assembly skills across different objects, regardless of their poses or motions. We emulate this ability by designing a problem space transformation~\cite{doshi2024problem} with SE(3)-equivariance, which maps all straight-line assembly motions in the world frame into equivalent top-down insertions in the task frame, allowing the RL agent to perceive them in a unified way. 

Formally speaking, given an assembly path $\tau$ with a pre-assembly position $p_d$ and the assembled position $p_a$, we define a path-centric transformation $\mathbf{T}$ conditioned on $\tau$ such that $\mathbf{T}(p_d) = (0,0,0)$ and $\mathbf{T}(p_d) - \mathbf{T}(p_a) = (0,0,\|p_d-p_a\|)$. Thus, the agent's observation is the transformed position of the part being assembled (under unknown noise), and its action is the transformed delta position, i.e., the ideal position change in the path-centric frame. 
Thanks to equivariance, the agent only needs to learn top-down insertions and can omit orientation from both observations and actions, simplifying the learning setup.
We use a task-space impedance (TSI) controller to enable smooth and compliant insertions, with impedance gains similarly transformed into the path-centric frame to maintain consistent behavior across different assembly directions. 
This design ensures that the observation and action spaces are minimal yet essential, facilitating generalizability among different assembly tasks, and is transferable to different robot arms and end effectors.

\subsection{Plan-Guided Residual Action} 

We find that guidance from the planned open-loop action helps learning by injecting prior knowledge about the coarse assembly direction. Thus, we adopt the idea of residual action~\cite{silver2018residual, ankile2024imitationrefinementresidual} in RL, where the policy outputs only the corrective action on top of the open-loop action, allowing the policy to focus on refining the assembly rather than learning the full assembly behavior. In practice, residual action warm-starts policy learning and typically leads to faster and better convergence.

\subsection{Minimalist Reward Design and Sim-to-Real Transfer} 

Surprisingly, our insertion reward is simply the negative L2 distance from the current part position to the goal position. This form is dense and sufficient enough for learning effective local insertion policies as our initial state is the pre-assembled pose given by the planner, which is already in the proximity of the goal thus does not require expensive exploration or complex reward engineering.

Due to the sim-to-real gap from misaligned dynamics, we adopt 1) domain randomization with 3mm-noised initialization on object pose during training and 2) Policy-Level Action Integrator (PLAI)~\cite{tang2023industreal} during deployment to ease the sim-to-real transfer of RL policies, which improves action consistency by incrementally applying policy outputs to the last desired state instead of the current state. 
PLAI applies policy actions as $s_{t+1}^{d} = s_{t}^{d} \oplus \Pi(o_t)$, where $s_{t}^{d}$ represents the desired state at time $t$, $\Pi(o_t)$ is the policy action computed based on the current observation $o_t$, and $\oplus$ denotes the composition operation, instead of the nominal approach $s_{t+1}^{d} = s_t \oplus \Pi(o_t)$ which is prone to error accumulation.

\section{Experiments}

\subsection{Benchmark Suite and Experimental Setup}

We develop a diverse benchmark suite spanning furniture, toys, and industrial equipment, which includes \textbf{beam} (5 parts), \textbf{plumber block} (5 parts), \textbf{car} (6 parts), \textbf{gamepad} (6 parts), \textbf{cooling manifold} (7 parts), \textbf{duct} (8 parts), and \textbf{stool} (9 parts). These assemblies cover various geometries and connection types found in real-world applications, with both top-down and sideway insertions, and are feasible for dual-arm robots with parallel grippers. 
For planning in simulation, we demonstrate on several different robots, including Franka Emika Panda, UFactory xArm7, and UR5e with different grippers.
We use Panda for systematic evaluations of policy training and real-world execution. See more details on the experimental setup and hyper-parameters in App.~\ref{app:setup}.

\subsection{Planning Multi-Step Assembly in Simulation}

{\bf Efficiency:} Table~\ref{tab:planning_speed} shows the breakdown of planning time by stages for different assemblies. Our overall speed is on the order of minutes to solve for optimal plans given efficient parallelization.

{\bf Optimality:} Table~\ref{tab:planning_score} shows the objective scores of optimized sequences surpassing the random ones by a large margin with priorities from $f_1$ to $f_4$ (see App.~\ref{app:seq_opt} for details on the score definitions).

{\bf Generality:} Please see App.~\ref{app:plan_demo} for visual demonstrations of planning with different robot arms (Panda, xArm7, UR5e) and grippers. Our planning framework is general to any given hardware.

\begin{table}[t]
\centering
\small
\begin{minipage}[t]{0.48\linewidth}
\caption{Planning runtime breakdown of each assembly. Stages marked with * are parallelized, while others have yet to be parallelized.}
\label{tab:planning_speed}
\centering
\begin{adjustbox}{width={\linewidth}}
\begin{tabular}{cc|cccccc}
\toprule
\multicolumn{2}{c|}{\multirow{2}{*}{\textbf{Assembly}}} & \multicolumn{6}{c}{\textbf{Runtime (s)}} \\ 
& & Prec$^*$ & Grasp$^*$ & Seq & Fixture & Motion & \textbf{Total} \\
\midrule
\multicolumn{2}{c|}{Beam} & 19.7 & 35.7 & 0.2 & 1.7 & 115.9 & 173.2 \\
\multicolumn{2}{c|}{Plumber} & 21.6 & 31.8 & 12.6 & 0.9 & 118.9 & 185.7 \\
\multicolumn{2}{c|}{Car} & 23.4 & 52.9 & 0.9 & 1.9 & 127.4 & 206.4 \\
\multicolumn{2}{c|}{Gamepad} & 22.2 & 37.2 & 4.5 & 3.6 & 117.4 & 184.8 \\
\multicolumn{2}{c|}{Manifold} & 20.9 & 162.7 & 5.2 & 1.9 & 149.6 & 340.4 \\
\multicolumn{2}{c|}{Duct} & 93.9 & 102.3 & 208.1 & 2.5 & 185.9 & 592.7 \\
\multicolumn{2}{c|}{Stool} & 57.2 & 109.1 & 8.0 & 4.0 & 324.5 & 502.7 \\
\bottomrule
\end{tabular}
\end{adjustbox}
\end{minipage}
\hfill
\begin{minipage}[t]{0.48\linewidth}
\caption{Objective comparisons between optimal and random sequences. Higher is better for $f_1$, $f_4$; lower is better for $f_2$, $f_3$.}
\label{tab:planning_score}
\centering
\begin{adjustbox}{width={\linewidth}}
\begin{tabular}{c|cccc}
\toprule
\multicolumn{1}{c|}{\multirow{2}{*}{\textbf{Assembly}}} & \multicolumn{4}{c}{\textbf{Objective Values (Optimal / Random)}} \\ 
& $f_1\uparrow$ & $f_2\downarrow$ & $f_3\downarrow$ & $f_4\uparrow$ \\
\midrule
Beam & 4.00 / 0.88 & 4.00 / 0.88 & 0.48 / 0.39 & 119.8 / 6.8 \\
Plumber & 4.00 / 1.27 & 2.00 / 1.79 & 0.18 / 0.33 & 293.0 / 93.4 \\
Car & 4.00 / 1.72 & 1.00 / 1.64 & 0.21 / 0.94 & 140.6 / 35.2 \\
Gamepad & 5.00 / 1.76 & 1.00 / 1.56 & 1.40 / 0.63 & 428.6 / 52.1 \\
Manifold & 6.00 / 2.62 & 1.00 / 2.17 & 0.18 / 0.69 & 12.5 / 18.7 \\
Duct & 6.00 / 3.00 & 1.00 / 2.88 & 0.09 / 0.39 & 536.0 / 101.3 \\
Stool & 8.00 / 3.43 & 6.00 / 3.24 & 0.03 / 0.54 & 322.2 / 87.9 \\
\bottomrule
\end{tabular}
\end{adjustbox}
\end{minipage}
\vspace{-2mm}
\end{table}

\subsection{Learning Single-Step Assembly in Simulation}

\begin{table*}[t]%
\small
\caption{\% of successful steps without intervention in simulation evaluations.}
\vspace{-4mm}
\label{tab:success_sim}
\begin{center}
\begin{adjustbox}{width={\textwidth}}
\begin{tabular}{c|ccccccc}

\toprule
\multirow{2}{*}{\textbf{Method}} & \multicolumn{7}{c}{\textbf{\% of Successful Steps without Intervention (Simulation)}} \\ 
& Beam & Plumber Block & Car & Gamepad & Cooling Manifold & Duct & Stool \\
\midrule
Open-Loop Tracking
& 21.48 & 24.22 & 2.34 & 2.34 & 3.91 & 18.75 & 0.00 \\
Part Specialist Policy (PS)
& \textbf{98.63} & 84.08 & \textbf{90.82} & \textbf{87.60} & \textbf{94.63} & \textbf{100.00} & \textbf{78.91} \\
Assembly Specialist Policy (AS)
& \textbf{99.12} & \textbf{97.46} & 70.12 & \textbf{88.87} & \textbf{95.02} & 96.58 & 76.66 \\
Assembly Generalist Policy (AG)
& \textbf{98.83} & 81.64 & 60.55 & 71.48 & 89.06 & 89.84 & 58.59 \\
\bottomrule

\end{tabular}
\end{adjustbox}
\end{center}
\vspace{-8mm}
\end{table*}%

We use Isaac Gym~\cite{makoviychuk2021isaac} for training RL policies and performing simulation evaluations, with the PPO code from RL Games~\cite{rl-games2021}.
Table~\ref{tab:success_sim} presents the average \% of successful steps for assembling our benchmark assemblies in simulation across 1024 random trials using different methods: 1) \textbf{Open-Loop Tracking}: A baseline that strictly follows the pre-planned path without feedback correction. 2) \textbf{Part Specialist Policy (PS)}: Policies trained on individual pairs of parts. 3) \textbf{Assembly Specialist Policy (AS)}: Policies trained on all parts within a single assembly. 4) \textbf{Assembly Generalist Policy (AG)}: Policies trained on all parts from all assemblies in our suite, aiming for broad generalization.

The results show that open-loop tracking exhibits the lowest success rates across all assemblies, indicating its limitations in handling uncertainties and variations.
The AS policy demonstrates competitive performance as the PS policy, suggesting that a shared policy across different parts in an assembly can generalize well. It may sacrifice some part-specific optimization, but can transfer the knowledge between similar parts.
The AG policy, while slightly less effective than the other RL counterparts, still demonstrates robust performance, suggesting that learning a single shared policy across different assemblies is promising, given the equivariant representations. Furthermore, the success rates vary across different assemblies, with simpler assemblies like the Beam and Cooling Manifold achieving higher performance across all methods, while more complex assemblies such as the Gamepad and Stool exhibit lower success rates due to their intricate geometries and constraints.

\subsection{Executing Multi-Step Assembly in Real World}

Table~\ref{tab:success_real_1} shows the \% of successful steps on benchmark assemblies evaluated in the real world (step-wise statistics), and Table~\ref{tab:success_real_2} shows the multi-step cumulative success rates with 0/1/2 total interventions for failure recovery (overall statistics). All numbers are averaged across three complete multi-step real experiments, which translate to thousands of total assembly steps. We deploy stochastic policies with state-based success detection and allow up to three trials per step until success. For qualitative results on real-world multi-step executions, please refer to Fig. 1 and App.~\ref{app:plan_demo}.

\textbf{Ours:} We use both AS and AG policies for real-world comparisons. For AG, we perform out-of-distribution (OOD) evaluations by training 7 generalist policies, where each one is trained on the other 6 benchmark assemblies (excluding the test assembly).
Remarkably, these OOD generalist policies still achieved comparable performance to specialist policies trained directly on the test assembly, which indicates that through our approach, insertion strategies learned from a diverse set of assemblies can effectively transfer to novel, unseen assemblies.

\textbf{Baseline:} Since Fabrica is the first to assemble general multi-part objects with only CAD input, identifying a comparable SOTA baseline is challenging.
The closest work is ASAP~\cite{tian2024asap}, which performs single-arm kinematic feasibility search without sequence/grasp optimization or closed-loop control. To compare, we adapted it by planning with dual-arm and adding our RL policy. 
Without optimized part sequencing and grasping, ASAP performed substantially worse, often struggling even with two interventions, which emphasizes the contributions of our planning optimizations.

\textbf{Ablation:} To understand how much optimizing part sequences and grasps helps, we conduct ablation studies on our method by removing those optimizations respectively. We observed that suboptimal sequencing often caused instability due to inadequate support of critical neighboring parts and more part drifts due to unnecessary re-grasps. Meanwhile, suboptimal grasp selection frequently caused part slippage due to insufficient contact area or inadequate resistance to external torques. 
Thus, our planner inherently accounts for control-level uncertainties, and results demonstrate that selecting effective part sequences and grasps significantly enhances assembly reliability. 
For control, we observed that our path-centric transformation is crucial for generalizing across varying assembly directions. Policies trained without it perform significantly worse when multiple directions are involved.
For more studies on the effects of path-centric transformation and residual actions introduced in Sec~\ref{sec:learning}, and success increase w.r.t. the number of trials per step, please see App.~\ref{app:ablation}.

\begin{wrapfigure}{h}{0.4\textwidth}
    \vspace{-5mm}
    \centering
    \includegraphics[width=0.4\textwidth]{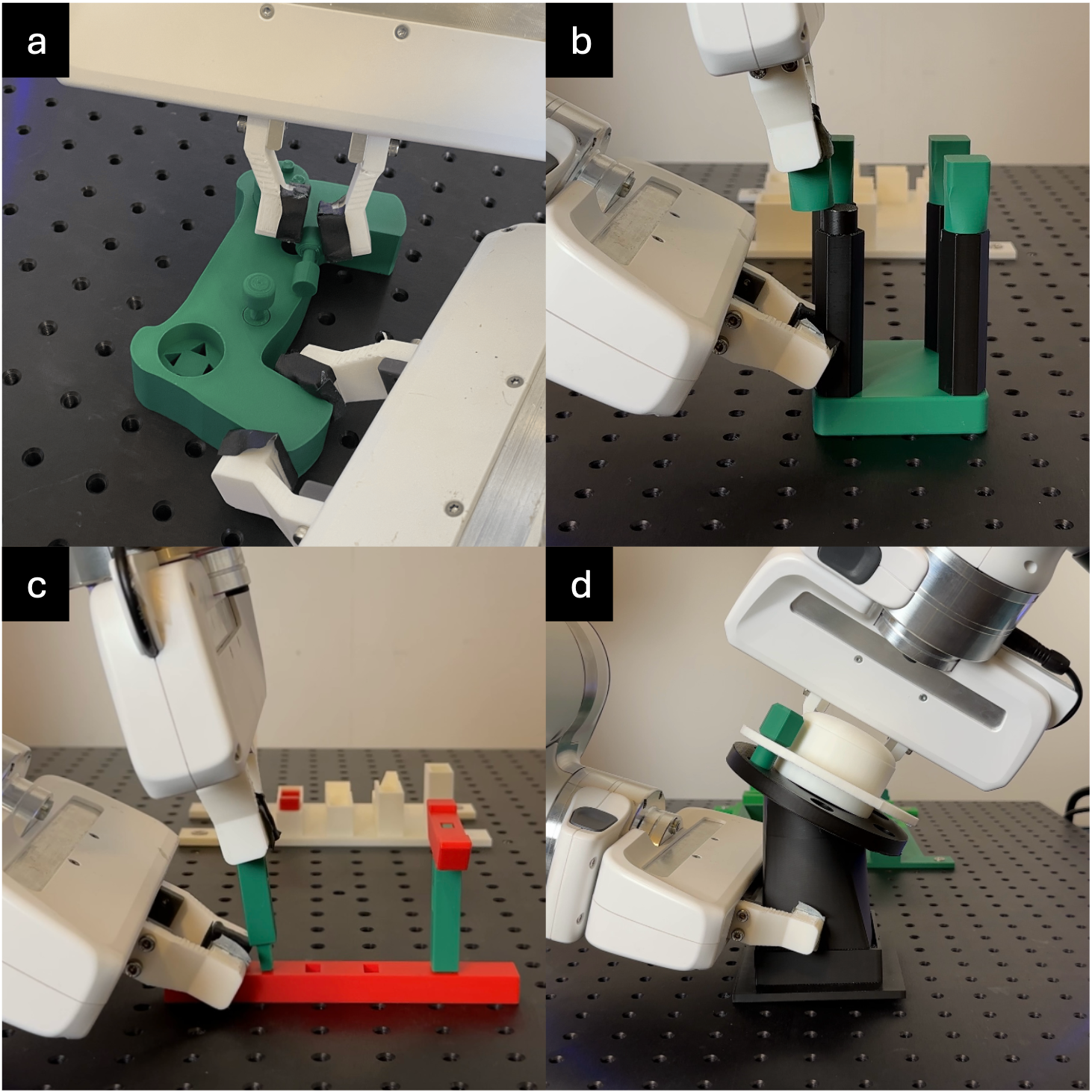}
    \label{fig:failure}
    \vspace{-5mm}
\end{wrapfigure}

\textbf{Failures:} We observed a noticeable gap between simulation and real-world performances. Thus, we present a detailed analysis of common failure cases shown in the right figure: a) Small parts slip between gripper pads during insertion attempts; b) Cumulative error accrued during the assembly of large assemblies increases the displacement error of final part insertions; c) The holding gripper is not modeled during RL training, causing unexpected part obstructions in the real world; d) Unstabilized parts shift previously assembled parts during insertion. We assume a 3mm noise in simulation given that the base part is stably held. However, many sources of real-world error lead to much more significant errors than simulated, which are non-trivial challenges for future work. 
Due to these failures, all methods achieve near-zero multi-step success rates without intervention due to inherently challenging steps causing consistent failures. However, with minimal interventions, our method significantly outperforms others, reaching 81\% success with one intervention and 95\% with two.

\begin{table*}[t]%
\small
\caption{\% of successful steps without intervention in real-world evaluations.}
\vspace{-4mm}
\label{tab:success_real_1}

\begin{center}
\begin{adjustbox}{width={\textwidth}}
\begin{tabular}{cc|ccccccc|c}
\toprule
\multicolumn{2}{c|}{\multirow{2}{*}{\textbf{Method}}} & \multicolumn{8}{c}{\textbf{\% of Successful Steps without Intervention (Real World)}} \\ 
& & Beam & Plumber Block & Car & Gamepad & Cooling Manifold & Duct & Stool & \textbf{Overall} \\
\midrule
\multirow{2}{*}{\shortstack{Ours}} & AS
& \textbf{75} & \textbf{83} & \underline{80} & \textbf{87} & \underline{72} & \underline{71} & \textbf{92} & \textbf{80} \\
& AG (OOD)
& \underline{67} & \underline{75} & \textbf{93} & \underline{80} & \underline{72} & \textbf{81} & \textbf{92} & \textbf{80} \\
\midrule
\multirow{2}{*}{\shortstack{Baseline}}  & ASAP (Adapted)
& 50 & 42 & 67 & 33 & 55 & 52 & \underline{75} & 55 \\
& Open-Loop Tracking
& 42 & 25 & 20 & 20 & 17 & 14 & 21 & 23 \\
\midrule
\multirow{3}{*}{\shortstack{Ablation}} 
& w/o Part Seq Optim
& \textbf{75} & \underline{75} & 73 & 40 & 61 & \underline{71} & \textbf{92} & \underline{71} \\
& w/o Grasp Optim
& 42 & \textbf{83} & 47 & 60 & 67 & 67 & \underline{75} & 64 \\
& w/o Path-Centric Transform
& \underline{67} & 67 & 20 & 53 & \textbf{78} & 38 & \textbf{92} & 61 \\
\bottomrule
\end{tabular}
\end{adjustbox}
\end{center}
\vspace{-2mm}

\end{table*}

\newcommand{\alignnum}[3]{\makebox[14pt][r]{#1}/\makebox[14pt][r]{#2}/\makebox[14pt][r]{#3}}

\begin{table*}[t]%
\small
\caption{Multi-step cumulative success rate with 0/1/2 interventions in real-world evaluations.}
\vspace{-4mm}
\label{tab:success_real_2}

\begin{center}
\begin{adjustbox}{width={\textwidth}}
\addtolength{\tabcolsep}{-0.5em}
\begin{tabular}{cc|ccccccc|c}
\toprule
\multicolumn{2}{c|}{\multirow{2}{*}{\textbf{Method}}} & \multicolumn{8}{c}{\textbf{Multi-Step Cumulative Success Rate with 0/1/2 Interventions (\%) (Real World)}} \\ 
& & Beam & Plumber Block & Car & Gamepad & Cooling Manifold & Duct & Stool & \textbf{Overall} \\
\midrule
\multirow{2}{*}{\shortstack{Ours}}  & AS
& \alignnum{0}{100}{100} & \alignnum{33}{100}{100} & \alignnum{0}{100}{100} & \alignnum{33}{100}{100} & \alignnum{0}{67}{67} & \alignnum{0}{0}{100} & \alignnum{33}{100}{100} & \alignnum{\textbf{15}}{\textbf{81}}{\underline{95}} \\
& AG (OOD)
& \alignnum{0}{67}{100} & \alignnum{0}{100}{100} & \alignnum{67}{100}{100} & \alignnum{0}{100}{100} & \alignnum{0}{33}{100} & \alignnum{0}{67}{100} & \alignnum{33}{100}{100} & \alignnum{\underline{10}}{\textbf{81}}{\textbf{100}} \\
\midrule
\multirow{2}{*}{\shortstack{Baseline}}  & ASAP (Adapted)
& \alignnum{0}{0}{100} & \alignnum{0}{0}{67} & \alignnum{0}{33}{100} & \alignnum{0}{0}{0} & \alignnum{0}{0}{33} & \alignnum{0}{0}{0} & \alignnum{0}{0}{100} & \alignnum{0}{5}{57} \\
& Open-Loop Tracking
& \alignnum{0}{0}{67} & \alignnum{0}{0}{33} & \alignnum{0}{0}{0} & \alignnum{0}{0}{0} & \alignnum{0}{0}{0} & \alignnum{0}{0}{0} & \alignnum{0}{0}{0} & \alignnum{0}{0}{14} \\
\midrule
\multirow{3}{*}{\shortstack{Ablation}} 
& w/o Part Seq Optim
& \alignnum{0}{100}{100} & \alignnum{33}{67}{100} & \alignnum{0}{67}{100} & \alignnum{0}{0}{0} & \alignnum{0}{0}{67} & \alignnum{0}{0}{100} & \alignnum{33}{100}{100} & \alignnum{\underline{10}}{48}{81} \\
& w/o Grasp Optim
& \alignnum{0}{0}{67} & \alignnum{33}{100}{100} & \alignnum{0}{0}{33} & \alignnum{0}{0}{100} & \alignnum{0}{0}{100} & \alignnum{0}{0}{67} & \alignnum{0}{0}{100} & \alignnum{5}{14}{81} \\
& w/o Path-Centric Transform
& \alignnum{0}{67}{100} & \alignnum{0}{67}{100} & \alignnum{0}{0}{0} & \alignnum{0}{0}{67} & \alignnum{0}{67}{100} & \alignnum{0}{0}{0} & \alignnum{33}{100}{100} & \alignnum{5}{43}{67} \\

\bottomrule
\end{tabular}
\end{adjustbox}
\end{center}
\vspace{-6mm}

\end{table*}

\section{Conclusion}
\label{sec:conclusion}

We presented Fabrica, a dual-arm robotic system that innovates and integrates global hierarchical planning with local generalist policy learning for autonomous multi-part assembly. To support reproducible and rigorous evaluation, we introduced a comprehensive benchmark suite covering diverse multi-part assemblies. Fabrica is the first to demonstrate robust and generalizable performance across a wide range of real-world assembly tasks. We discuss limitations and future work in Sec~\ref{sec:limitation}.


\clearpage
\section{Limitations and Future Work}
\label{sec:limitation}

While Fabrica shows promising results for autonomous multipart assembly, there remain several limitations and opportunities for future extension. 

\textbf{Assumptions:} The assumptions we make in this problem formulation are the following:
\begin{enumerate}[label=\assumpitem{A}{{\arabic*}}]
    \item {\it Insertion-only assembly}: We assume that the mating between two parts only involves an insertion motion, without requiring skills like screwing or sliding.  \label{assump:insert-only}  
    \item {\it No subassembly reorientation}: The final assembled positions of all parts are assumed to be given and fixed during the assembly process. This means that no further movement or re-orientation is allowed once a part is assembled.\label{assump:fixed-goal-pose}
    \item {\it Monotonic assembly}: Each part is only moved once, without considering regrasps, in-hand manipulation, or handovers.\label{assump:monotonic}
    \item {\it No force and torque constraints for the robots}: We assume all parts are light compared to the robot payload.\label{assump:no-force}
    \item {\it A finite grasp set for each part}: $\forall g \in \sigma, g \in \mathcal{G}[o], |\mathcal{G}[o]| = N$, where $\mathcal{G}[o]$ can be computed by any grasp generator.\label{assump:finite_grasp}
\end{enumerate}
The above assumptions leave areas such as handling heavier parts, managing grasp slippage, and performing other operations such as screwing or sliding unaddressed. 
Incorporating these capabilities would significantly improve the robustness and applicability of Fabrica in more complex and diverse assembly tasks.

\textbf{Dexterity:} Moreover, the current setup enforces a fixed part pose once assembled.
This is in contrast with the more dexterous human assembly behavior, where one would constantly reorient the partial assembly so that the parts are easily reachable.
However, allowing reorientation would introduce additional planning overhead and more uncertainty for control due to potential subassembly instability.
Addressing it will enable a more dexterous robotic system that can handle large assemblies that are beyond the reach of the current system, e.g., a large tabletop with parts on both sides.

\textbf{Hardware capability:} Compared to the existing multipart assembly dataset with thousands of objects \cite{tian2022assemble}, our current benchmark is limited in its size and diversity.
This is because we want to ensure that the benchmark tasks are achievable by commonly used parallel grippers, but these grippers have a limited grasp width and thus cannot establish stable antipodal grasps for parts with large and complex geometries.
However, to broaden the assembly capability, we can envision either a multi-finger hand or a multi-tool system in which the robots can switch specialized grippers according to the part geometry, and our planning and control system could be adapted to this setting.

\textbf{Perception:} Integrating vision systems for alignment feedback could greatly improve the accuracy and adaptability of the assembly process. By incorporating perception, the system could enable direct bin-picking, allowing it to grasp parts from random, unknown initial poses instead of requiring a specialized pickup fixture. 
However, bin-picking remains a well-known challenge in industry, particularly in terms of robustness and generalizability across arbitrary part geometries and physical properties. Addressing these challenges requires substantial research and development efforts, but would significantly expand the practical applications of our approach.

\textbf{Data collection:} Collecting real-world assembly data is challenging due to the task’s long-horizon, contact-rich nature and the high cost of acquiring or fabricating diverse assembly assets. Fabrica addresses these barriers by enabling fully autonomous data collection in simulation and requiring only minimal human intervention in the real world. In future work, we aim to leverage this capability to generate large-scale, diverse datasets of assembly trajectories. These datasets can facilitate broader research in generalizable policy learning, sim-to-real transfer, and foundation models for robotic manipulation. Moreover, because assembly is one of the most constrained and demanding manipulation tasks, learning from assembly data has the potential to positively transfer to a wide range of general-purpose manipulation skills.

\clearpage
\acknowledgments{This work is funded by Autodesk, and in part by NSF 1846368 \& 2313076. Yijiang Huang is supported by the SNSF Ambizione program. The authors would like to thank Bingjie Tang and Lars Ankile for insightful discussions during the RL environment setup, Xiang Zhang and Yotto Koga for valuable advice on the early physical setup, the members of Ted Adelson's lab at MIT for their support with the physical Panda robot, and the MIT SuperCloud and Lincoln Laboratory for HPC resources.}


\bibliography{reference}  

\clearpage
\appendix
\begin{appendices}
\setcounter{figure}{0}
\setcounter{table}{0}
\setcounter{footnote}{0}
\section{Qualitative Results}
\label{app:plan_demo}

\begin{figure*}[h]
    \centering
    \includegraphics[width=\textwidth]{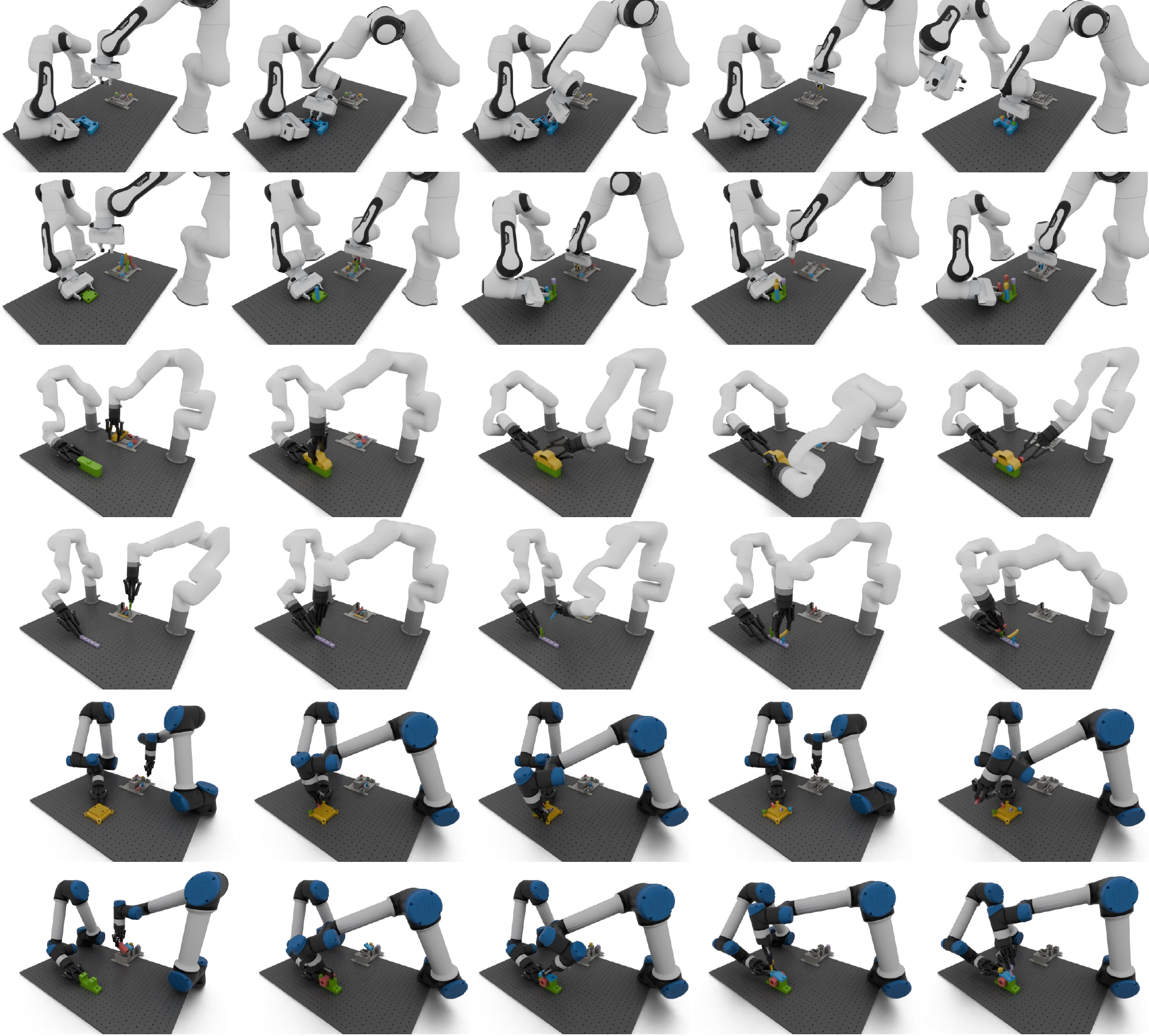}
    \caption{Step-by-step rendered assembly executions on different assemblies with different robots.}
    \label{fig:planning_demo}
\end{figure*}

\begin{figure*}[h]
    \centering
    \includegraphics[width=\textwidth]{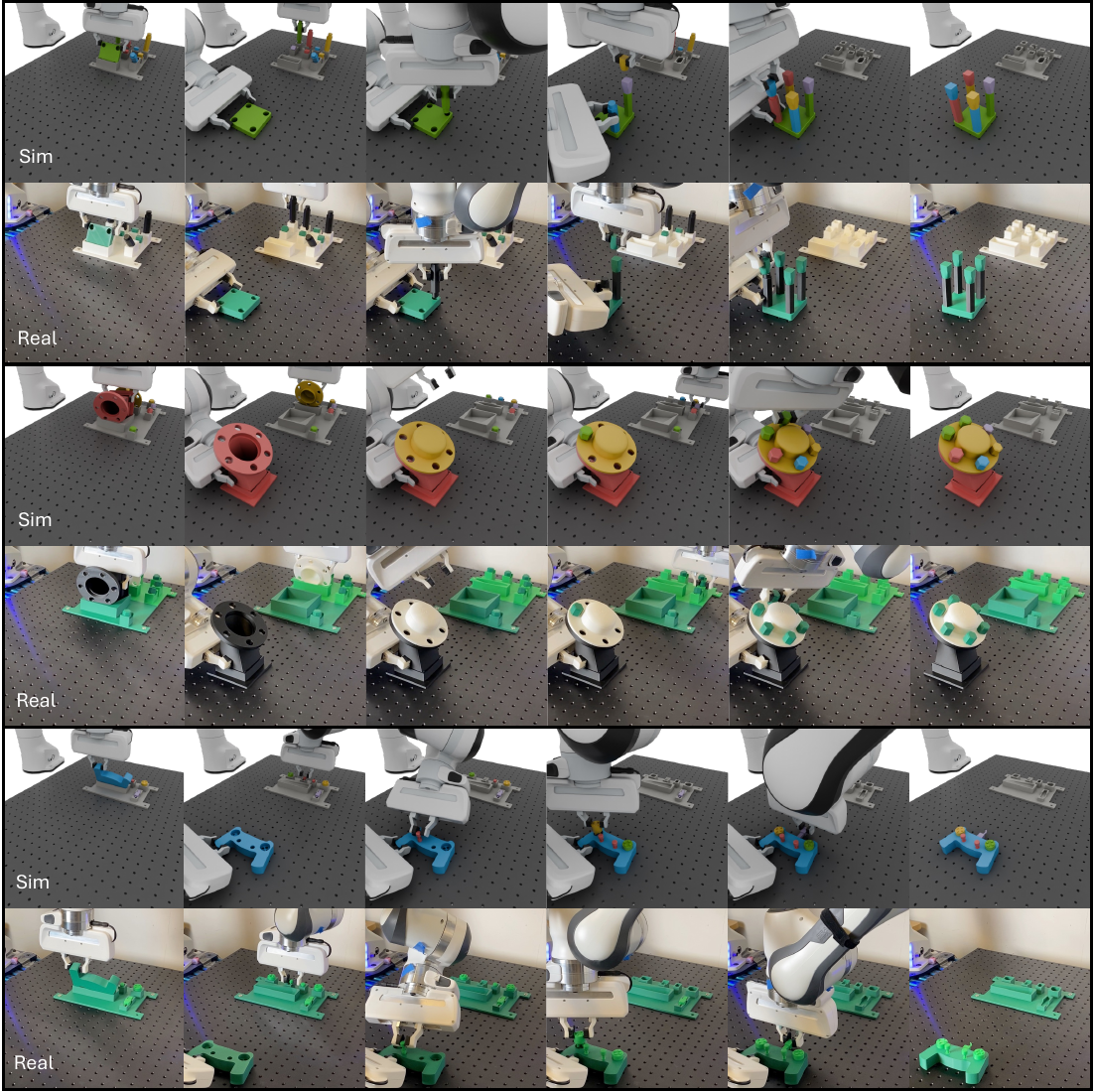}
    \caption{Step-by-step real-world assembly executions on different assemblies with Panda robots, with side-by-side correspondences between simulation and real.}
    \label{fig:full_real_demo}
\end{figure*}

Fig.~\ref{fig:planning_demo} and Fig.~\ref{fig:full_real_demo} show qualitative results of our system assembling a variety of multi-part objects using different robot arms, both in simulation and in the real world. Once the hardware configuration is provided, our planner works seamlessly with a wide range of robot arms and end-effectors without requiring additional tuning.

The assembly process begins with picking up parts from the fixture, then transferring them to the assembly area for either holding or insertion. The system determines when to switch roles between the two arms, for example by handing over a part or changing the holding grasp, in order to maintain stability and accuracy. After each insertion, the robot returns to retrieve the next part, and this process continues until the assembly is complete. Finally, both arms return to their initial positions.

All plans are generated automatically, including grasp selection, arm coordination, motion generation, and fixture design. The only manual input required is the initial setup of the workcell and placement of the hardware. 

For real-world execution, we demonstrate robust sim-to-real transfer across long-horizon assembly tasks, with consistent step-by-step correspondence between simulation and physical execution. The system maintains geometric and temporal alignment across all stages of the assembly, including grasping, insertion, and part switching. This highlights the reliability of our simulation-informed planning and policy execution.

\section{Problem Formulation}
\label{app:plan_formulation}

As in Sec.~\ref{sec:plan}, we formulate planning as optimizing assembly-hold sequences \(\phi\), grasps \(\sigma\), and robot motions \(\pi\):
\begin{equation}
\label{eq:optim}
\min_{\phi, \sigma, \pi} E\big( \Phi_{i=1}^{n}\vec{f}(\phi_{1:i}, \sigma_{i-1:i}, \pi_{i})\big) \quad
\text{s.t.} \quad C_{\text{prec}}(\phi) \leq 0, \hspace{0.2em}
C_{\text{kin}}(\phi, \sigma, \pi) = 0, \hspace{0.2em}
C_{\text{col}}(\phi, \sigma, \pi) \leq 0
\end{equation}
A more detailed breakdown of constraints $C_{\text{prec}}(\phi)$, $C_{\text{kin}}(\phi)$, $C_{\text{col}}(\phi)$ is shown below.
\begin{subequations}
 \begin{align}
        \min_{\phi, \sigma, \pi} \ &\vec{F}(\phi,\sigma) \nonumber\\
        \textrm{s.t.} \quad &\forall i \in [1, n], \nonumber\\
        &C_\text{prec}(\phi_{1:i}) \leq 0 \label{constr:prec}\\
        &C_\text{kin}(o_{a,i},g_{a,i},\tau_{a,i}^g(0),p^0_{o_i}) = 0 \label{constr:kin-1}\\
        &C_\text{kin}(o_{a,i},g_{a,i},\tau_{a,i}^f(1),p^G_{o_i}) = 0 \label{constr:kin-2}\\
        &C_\text{kin}(o_{h,i},g_{h,i},\tau_{h,i}^f(1),p^G_{o_i}) = 0 \label{constr:kin-3}\\
        &C_\text{col}(\phi_{1:i},\pi[a,i],\pi[h,i]) \leq 0 \label{constr:collision}
 \end{align}
    \label{eq:planning_opt}%
\end{subequations}
where each motion $\tau: [0,1] \rightarrow \mathcal{Q}$ is a time-parametrized joint trajectory in the robot's configuration space $\mathcal{Q}$. \cref{constr:prec} are the precedence constraints that ensure the partially assembled parts are connected and do not collapse under gravity.
\cref{constr:kin-1,constr:kin-2,constr:kin-3} ensure the robot configuration, grasp, and the grasped object pose are kinematically consistent when picking, assembling, and holding.
\cref{constr:collision} ensures that both robots' trajectories do not collide with previously assembled parts and other static obstacles.

The key planning stages in Sec.~\ref{sec:plan} are further summarized here.
\begin{itemize}
    \item \Cref{sec:precedence}: Starting from a multi-part mesh model, we construct a precedence graph representing a minimum constraint set for any feasible sequence, considering only collision among parts.
    \item \Cref{sec:grasp}: To reduce online collision checking overhead during search, we pre-compute a discretized, collision-free grasp set for assembling and holding for all part pairs. Each feasible grasp is associated with a corresponding robot trajectory.
    \item \Cref{sec:seq_opt}: We leverage the precedence graph and precomputed grasp pairs to expand a state tree that contains all feasible part-grasp sequences and then search for an optimal part-grasp sequence that minimizes a grasp stability cost.
    \item \Cref{sec:fixture}: After the grasps are determined, we develop an automatic design algorithm to generate a pickup fixture, so that the planned grasp can be achieved easily without the need of a re-grasp.
    \item \Cref{sec:motion_plan}: With the assembly sequence fixed and all robot configurations determined for kinematic switches, i.e., pick-up, assembly, and hold, we plan for all transit and transfer motions.
\end{itemize}

\section{Algorithmic Details}
\label{app:algo}

We now present detailed mathematical formulations of each algorithm introduced in Sec.~\ref{sec:plan}.

\subsection{Part Precedence Planning}
\label{app:precedence}

\Cref{alg:precedence} and the following paragraphs provide the details of the part precedence planning algorithm.

\paragraph{Precedence Tier Generation}
Initially, an empty list of tiers \( T_\text{prec} \) is initialized. We use $O_r$ to represent all the remaining assembled parts, which starts from all parts \( O \). Although \( O_r \) is not empty, the algorithm constructs a new tier \( t_\text{prec} \) by evaluating each part \( o_i \in O_r \) to determine the feasibility of disassembly. Using motion planning, a disassembly path \( q_i \) is computed for \( o_i \). 
Specifically, we apply Assemble-Them-All~\cite{tian2022assemble}, a physics-based method for efficient disassembly motion planning given the highly constrained search space, and determine success based on a given timeout. 
If \( o_i \) can be disassembled, the pair \( (o_i, q_i) \) is added to \( t_\text{prec} \).
Once all feasible parts are processed, the constructed tier \( t_\text{prec} \) is added to \( T_\text{prec} \), and all \( o_i \in t_\text{prec}\) is removed from \( O_r \).
This process repeats until all parts are assigned to tiers.

\paragraph{Precedence Graph Generation}

The graph construction phase generates a directed graph \( G_\text{prec} \) that encodes precedence constraints among parts. An empty set \( O_e \) is initialized to track parts in earlier tiers (i.e., parts that are supposed to be disassembled earlier). For each tier \( t_\text{prec} \in T_\text{prec} \), the algorithm processes every part \( o_i \in t_\text{prec} \) by checking its disassembly path \( \tau_{o_i} \) for collisions with parts in earlier tiers \( O_e \). A collision check function identifies the set of colliding parts \( O_c \subseteq O_e \). For each \( o_c \in O_c \), an edge \( (o_i, o_c) \) is added to \( G_\text{prec} \), indicating that \( o_i \)'s disassembly depends on \( o_c \)'s prior removal. After processing all parts in \( t_\text{prec} \), the disassembled parts \( O_t \) are added to \( O_e \). The algorithm proceeds until all parts in all precedence tiers are added to the precedence graph.

\begin{algorithm}
\caption{Part Precedence Planning}
\label{alg:precedence}
\begin{algorithmic}[1]
\State \textbf{Input:} All parts $O$ with goal poses $p_O^G$
\State \textbf{Output:} Directed graph $G_\text{prec}$ representing precedence constraints

\State Initialize an empty list of tiers: $T_\text{prec} \gets \texttt{[]}$
\State Initialize an empty directed graph: $G_\text{prec} \gets \text{DiGraph}()$
\State Initialize all remaining parts $O_r \gets O$

\While{$O_r \neq \emptyset$} \Comment{Tier generation}
    \State Initialize an empty tier: $t_\text{prec} \gets \{\}$
    \For{each part $o_i \in O_r$}
        \State $\tau_{o_i} \gets \text{DisassemblyPath}(o_i, O_r)$
        \If{feasible $\tau_{o_i}$ is found}
            \State $t_\text{prec} \gets t_\text{prec} \cup \{(o_i, \tau_{o_i})\}$
        \EndIf
    \EndFor
    \State $O_r \gets O_r \setminus \{o_i \mid (o_i, \tau_{o_i}) \in t_\text{prec} \}$
    \State $T_\text{prec}.\text{Append}(t_\text{prec})$
\EndWhile

\State Initialize an empty set of parts in earlier tiers: $O_e \gets \emptyset$
\For{each tier $t_\text{prec} \in T_\text{prec}$} \Comment{Graph construction}
    \State Let $O_t \gets \{o_i \mid (o_i, \tau_{o_i}) \in t_\text{prec}\}$ \Comment{Parts in tier $t_\text{prec}$}
    \For{each $(o_i, \tau_{o_i}) \in t_\text{prec}$}
        \State $G_\text{prec}.\text{AddNode}(o_i, \text{path}=\tau_{o_i})$
        \State $O_c \gets \text{CheckPathCollision}(o_i, \tau_{o_i}, O_e)$ \Comment{$O_c$: colliding parts in $O_e$}
        \For{each $o_c \in O_c$}
            \State $G_\text{prec}.\text{AddEdge}(o_i, o_c)$
        \EndFor
    \EndFor
    \State $O_e \gets O_\text{e} \cup O_t$ \Comment{Update parts in earlier tiers}
\EndFor
\State \Return $G_\text{prec}$

\end{algorithmic}
\end{algorithm}

\subsection{Dual-Arm Grasp Filtering}
\label{app:grasp}

This section provides a detailed description of the sub-steps involved in the grasp filtering algorithm.

\paragraph{Single-Pose Grasp Feasibility Check}

\cref{alg:check_grasp_feasibility} evaluates whether a specific grasp configuration \( g \) is feasible for a target part \( o \) in its current pose \( p_o \). The algorithm first determines the set of preceding parts \( O_\text{prec} \) from the precedence graph \( G_\text{prec} \). 
For each gripper aperture (\( a_\text{grasp} \) and \( a_\text{release} \)), collision checks are performed involving the robot body, the gripper, the target part \( o \), the preceding parts \( O_\text{prec} \), and the environment obstacles \( E \). 
If any collision occurs between them, the grasp $g$ is not feasible. Additional collision checks are performed between the gripper and other non-preceding parts \( O_\text{all} \setminus O_\text{prec} \), and collision information is added to the grasp. Such collisions are not hard constraints because the non-preceding parts may get assembled later than the target part depending on the specific assembly sequence. In this case, the collision does not matter, but the information has to be recorded to find collision-free assembly sequences in the later stage.

\begin{algorithm}
\caption{Single-Pose Grasp Feasibility Check}
\label{alg:check_grasp_feasibility}
\begin{algorithmic}[1]
\State \textbf{Input:} Grasp $g$, robot $R$, target part $o$ with pose $p_o$, all parts $o \in O$ with their corresponding goal poses $p_o^G$, precedence graph $G_\text{prec}$
\State \textbf{Output:} Feasibility (\textbf{True}/\textbf{False}), updated grasp $g$ with collision and IK information

\State $O_\text{prec} \gets G_\text{prec}.\text{PrecedingParts}(o)$
\State $E \gets$ environment obstacles

\State $q_g^{R} \gets \text{IK}(R, g, o, p_o)$ \Comment{IK for robot to grasp part $o$ under pose $p_o$ with grasp $g$}
\If{feasible $q_g^{R}$ is found}
    \State Set robot $R$ configuration to $q_g^{R}$
    \For{\textbf{each} gripper aperture $a \in \{a_\text{grasp}, a_\text{release}\}$}
        \State Set gripper with aperture $a$
        \If{$\text{CheckCollision}(R, o, O_\text{prec}, E)$}
            \State \Return \textbf{False}, $g$
        \EndIf
        \State $\text{CheckCollision}(R, O \setminus O_\text{prec})$
    \EndFor
    \State Record collision and IK information to $g$
    \State \Return \textbf{True}, $g$
\Else
    \State \Return \textbf{False}, $g$
\EndIf

\end{algorithmic}
\end{algorithm}

\paragraph{Assembling Grasp Feasibility Check}

\cref{alg:check_grasp_assembling_feasibility} determines the feasibility of using a grasp \( g \) to disassemble a target part \( o \) along its disassembly path \( \tau_o \), derived from \( G_\text{prec} \). For each pose \( p_{o,t} \) along \( \tau_o \), the grasp is transformed accordingly and its feasibility is validated using \cref{alg:check_grasp_feasibility}. The aggregated collision and IK information across all poses is stored in \( g \). The grasp is feasible if all poses along \( \tau_o \) are validated.

\begin{algorithm}
\caption{Assembling Grasp Feasibility Check}
\label{alg:check_grasp_assembling_feasibility}
\begin{algorithmic}[1]
\State \textbf{Input:} Grasp $g$, robot $R$, target part $o$, all parts $O$ with goal poses $p_O^G$, precedence graph $G_\text{prec}$
\State \textbf{Output:} Feasibility (\textbf{True}/\textbf{False}), updated grasp $g$ with collision and IK information

\State $\tau_o \gets G_\text{prec}.\text{GetPath}(o)$ \Comment{Disassembly path of part $o$}

\For{\textbf{each} part pose $p_{o,t} \in \tau_o$} \Comment{Disassembling $o$ while grasping $o$}
    \State Transform grasp $g$ according to $p_o^t$ to obtain $g_t$
    \State feas, $g_t \gets$ CheckGraspFeas($g_t$, $o$, $p_{o,t}$, ...) \Comment{Alg. \ref{alg:check_grasp_feasibility}}
    \If{\textbf{not} feas}
        \State \Return \textbf{False}, $g$
    \EndIf
    \State Gather collision and IK information from $g_t$ to $g$
\EndFor

\State \Return \textbf{True}, $g$
\end{algorithmic}
\end{algorithm}

\paragraph{Holding Grasp Feasibility Check}

\cref{alg:check_grasp_holding_feasibility} evaluates whether a grasp \( g \) can securely hold a part \( o \) while allowing other parts \( o_i \) to be disassembled. The feasibility of \( g \) is first validated using \cref{alg:check_grasp_feasibility}. 

\begin{algorithm}
\caption{Holding Grasp Feasibility Check}
\label{alg:check_grasp_holding_feasibility}

\begin{algorithmic}[1]
\State \textbf{Input:} Grasp $g$, robot $R$, target part $o$, all parts $O$ with goal poses $p_O^G$, precedence graph $G_\text{prec}$
\State \textbf{Output:} Feasibility (\textbf{True}/\textbf{False}), updated grasp $g$ with collision and IK information

\State feas, $g \gets$ CheckGraspFeas($g$, $o$, $p_o^G$, ...) \Comment{Alg. \ref{alg:check_grasp_feasibility}}

\If{\textbf{not} feas}
    \State \Return \textbf{False}, $g$
\EndIf

\For{\textbf{each} part $o_i \in O_\text{all} \setminus \{o\}$}
    \State $q_{o_i} \gets G_\text{prec}.\text{GetPath}(o_i)$ \Comment{Disassembly path of $o_i$}
    \For{\textbf{each} part pose $p_{o_i,t} \in q_{o_i}$} \Comment{Disassembling $o_i$ while grasping $o$}
        \State CheckCollision($R$, $o_i$) and gather collision information to $g$
    \EndFor
\EndFor

\State \Return \textbf{True}, $g$
\end{algorithmic}
\end{algorithm}

\paragraph{Grasp Pair Filtering}

Putting the assembling and holding feasibility checks together, \cref{alg:grasp_plan} generates and filters the dual-arm grasp pairs. 
For each part \( o_i \), a set of candidate grasp poses \( \{g_k\}_{k=1}^{N_g} \) is generated. 
Each grasp is evaluated for assembling and holding feasibility using \cref{alg:check_grasp_assembling_feasibility,alg:check_grasp_holding_feasibility}, respectively, and feasible grasps are stored in \( \mathcal{G}^a[o_i] \) and \( \mathcal{G}^h[o_i] \).
Finally, iterating through all assembly-hold part pairs, the set of feasible assembly-hold grasp pairs \( \mathcal{G}^{a \times h} [o_a, o_h]\) only contains those that do not lead to collisions between the two robots.

\begin{algorithm}
\caption{Dual-Arm Grasp Pair Filtering}
\label{alg:grasp_plan}
\begin{algorithmic}[1]
\State \textbf{Input:} All parts $O$ with goal poses $p_O^G$, robots $R_a, R_h$, precedence graph $G_\text{prec}$
\State \textbf{Output:} Assembling grasps $\mathcal{G}^a$, holding grasps $\mathcal{G}^h$, assembling-holding grasp pairs $\mathcal{G}^{a \times h}$

\State Initialize empty dictionaries $\mathcal{G}^a, \mathcal{G}^h, \mathcal{G}^{a \times h} \gets \{:\}, \{:\}, \{:\}$

\For{\textbf{each} part $o_i$ in $O$} \Comment{Feasible grasp generation}
    \State $\mathcal{G}^a[o_i], \mathcal{G}^h[o_i] \gets \{\}, \{\}$
    \State Generate $N_g$ grasp poses $\{g_k\}_{k=1}^{N_g}$ on part $o_i$
    \For{\textbf{each} grasp pose $g_k$}
        \State $\text{feas}_a, g_a \gets \text{CheckAssemGraspFeas}(g_k, R_a, o_i, ...)$
        \If{$\text{feas}_a$}
            \State $\mathcal{G}^a[o_i] \gets \mathcal{G}^a[o_i] \cup \{g_a\}$
        \EndIf
        \State $\text{feas}_h, g_h \gets \text{CheckHoldGraspFeas}(g_k, R_h, o_i, ...)$
        \If{$\text{feas}_h$}
            \State $\mathcal{G}^h[o_i] \gets \mathcal{G}^h[o_i] \cup \{g_h\}$
        \EndIf
    \EndFor
\EndFor

\For{\textbf{each} part $o_a \in O$} \Comment{Feasible grasp pair filtering}
    \For{\textbf{each} part $o_h \in O$}
        \If{$o_a \in G_\text{prec}.\text{PrecedingParts}(o_h)$}
            \State \textbf{continue}
        \EndIf
        $\mathcal{G}^{a \times h}[(o_a,o_h)] \gets \{\}$
        \For{\textbf{each} grasp $g_a \in \mathcal{G}^a[o_a]$}
            \For{\textbf{each} grasp $g_h \in \mathcal{G}^h[o_h]$}
                \State Set robot $R_a$ configuration to $q_{g_a}^{R_a}$
                \State Set robot $R_h$ configuration to $q_{g_h}^{R_h}$
                \If{\textbf{not} CheckCollision$(R_a, R_h)$}
                    \State $\mathcal{G}^{a \times h}[(o_a,o_h)] \gets \mathcal{G}^{a \times h}[(o_a,o_h)] \cup 
                    \{(g_a, g_h)\}$
                \EndIf
            \EndFor
        \EndFor
    \EndFor
\EndFor

\State \Return $\mathcal{G}^a, \mathcal{G}^h, \mathcal{G}^{a \times h}$
\end{algorithmic}
\end{algorithm}

\subsection{Dual-arm Sequence-Grasp Optimization}
\label{app:seq_opt}

\Cref{alg:seq_opt} provides the pseudocode for the sequence-grasp optimization algorithm, followed by the formulas used to evaluate the transition edge cost.

\begin{algorithm}
\caption{Dual-Arm Assembly Sequence Planning}
\label{alg:seq_opt}
\begin{algorithmic}[1]
\State \textbf{Input:} All parts $O$, precedence graph $G_\text{prec}$, assembling grasps $\mathcal{G}^a$, assembling-holding grasp pairs $\mathcal{G}^{a \times h}$
\State \textbf{Output:} Optimal assembly part-grasp sequence $S^*$
\State Initialize an empty tree $T_G \gets$ DiGraph()
\State S $\gets$ [] \Comment{Initialize an empty search stack.}
\For{\textbf{each} $o \in O$}
    \If{$G_\text{prec}\text{.SucceedingParts}(o) = \emptyset$} 
        \For{\textbf{each} $g \in \mathcal{G}^a[o]$}
            \State $T_G$.AddNode(($a, O\setminus\{o\}, o, g$)) \Comment{Root nodes}
            \State \textbf{push}(S, ($a, O\setminus\{o\}, o, g$))
        \EndFor
    \EndIf
\EndFor

\While{S \textbf{not} empty}
    \State $(t_i, O_i, o_i, g_i) \gets$ \textbf{pop}(S)
    \State $t_{i+1} \gets h$ \textbf{if} $t_i = a$ \textbf{else} $t_{i+1} \gets a$
    \For{\textbf{each} $o_{i+1} \in O_i$}
        \State $O_{i+1} \gets O_i \setminus \{ o_i \}$ \textbf{if} $t_{i+1} = a$ \textbf{else} $O_{i+1} \gets O_i$
        
        \If{$(O_i \cup \{o_i\}) \cap G_\text{prec}\text{.PrecedingParts}(o_{i+1}) \neq \emptyset$}
            \State \textbf{continue}  \Comment{Precedence check}
        \EndIf
        
        \For{\textbf{each} $g_{i+1} \in \mathcal{G}^{t_{i+1}}[o_{i+1}]$}
                \State $O_c \gets \mathcal{G}^{t_{i+1}}[o_{i+1}][g_{i+1}].\text{CollidingSet}$ \Comment{Precomputed collision set}
                \If{$O_c \cap O_{i+1} \neq \emptyset$}
                    \State \textbf{continue} \Comment{State collison check}
                \EndIf
     
                \If{$(g_{i+1}, g_i) \in \mathcal{G}^{t_{i+1}\times t_i}[(o_{i+1},o_i)]$}
                    \State $T_G$.AddEdge((($t_{i}, O_{i}, o_{i}, g_{i}$), ($t_{i+1}, O_{i+1}, o_{i+1}, g_{i+1}$))) \Comment{Grasp pair validity check}
                    \State \textbf{push}(S, ($t_{i+1}, O_{i+1}, o_{i+1}, g_{i+1}$))
                \EndIf
        \EndFor
    \EndFor
\EndWhile

\For{\textbf{each} $e_i \in T_G$.Edges()} \Comment{Objective evaluation}
    \State $((t_i, O_i, o_i, g_i), (t_j, O_j, o_j, g_j)) \gets e_i$
    \If{$t_i = a, t_j = h$} \Comment{Assembling-holding step}
        \State $e_i.\vec{f} \gets \vec{f}(O_i, o_i, g_i, o_j, g_j)$
    \Else \Comment{Holding transition step}
        \State $e_i.\vec{f} \gets \vec{0}$
    \EndIf
\EndFor

\State $S^* \gets [o_{a,1}^*, g_{a,1}^*, o_{h,2}^*, g_{h,2}^*, ..., o_{a,n}^*, g_{a,n}^*] \gets \text{DP}(T_G, \Phi)$ \Comment{Dynamic programming}
\State \Return $S^*$
\end{algorithmic}
\end{algorithm}

\paragraph{Objective evaluation}

\begin{itemize}
    \item Maximizing the number of supportive parts held ($f_1$): Part A is supportive to part B if A is in the preceding parts of B in $G_\text{prec}$.
    \item Minimizing the number of holding grasp transitions ($f_2$): Holding grasp transitions can be counted simply by comparing whether the holding grasps in consecutive steps are the same.
    \item Minimizing approximated torque for assembling grasps ($f_3$): 
    \begin{equation}
    \frac{\left\| \boldsymbol{\tau}_{\text{part}} + \boldsymbol{\tau}_{\text{grasp}} \right\|}{N_{\text{grasp}}}
    \end{equation}
    Where:
    \begin{align*}
    \boldsymbol{\tau}_{\text{part}} &= \sum_{i=1}^{N_{\text{part}}} \left( \mathbf{r}_{i} - \mathbf{c}_{\text{part}} \right) \times \frac{\mathbf{d}_{\text{contact}}}{N_{\text{part}}} \\\\
    \boldsymbol{\tau}_{\text{grasp}} &= \sum_{j=1}^{N_{\text{grasp}}} \left( \mathbf{r}_{j} - \mathbf{c}_{\text{part}} \right) \times \frac{-\mathbf{d}_{\text{contact}}}{N_{\text{grasp}}}
    \end{align*}
    Where:
    \begin{itemize}
        \item $\mathbf{r}_{i}, \mathbf{r}_{j}$ are the position vectors of contact points on the part and grasp, respectively.
        \item $\mathbf{c}_{\text{part}}$ is the center of mass of the part.
        \item $\mathbf{d}_{\text{contact}}$ is the contact direction vector.
        \item $N_{\text{grasp}}$ and $N_{\text{part}}$ are the number of contact points for the grasp and part, respectively.
    \end{itemize}
    \item Maximizing part contact area for holding grasps, weighted by the orientation difference from the paired assembling grasps ($f_4$): 
    \begin{equation}
    \Delta \theta_{\text{rotation}} \times N_{\text{hold}}
    \end{equation}
    Where:
    \begin{align*}
    \Delta \theta_{\text{rotation}} &= \left\| \mathbf{R}_{\text{hold}}^{-1} \cdot \mathbf{R}_{\text{assemble}} \right\|
    \end{align*}
    Where:
    \begin{itemize}
        \item $\Delta \theta_{\text{rotation}}$ represents the angular difference between the holding and assembling grasps.
        \item $\mathbf{R}_{\text{hold}}$ and $\mathbf{R}_{\text{assemble}}$ are the rotation matrices derived from the quaternions of the holding and assembling grasps, respectively.
        \item $N_{\text{hold}}$ is the number of contact points in the holding grasp.
    \end{itemize}
\end{itemize}

\subsection{Grasp-Oriented Pickup Fixture Generation}\label{app:fixture}

The fixture generation process begins with the computation of an appropriate pickup pose for each part. 
Our goal is to enforce a top-down pickup grasp without requiring re-grasping during the transition from pickup to assembly. 
Therefore, we can derive the pickup orientation of each part given the final assembled pose of them and the optimal grasp planned at the assembled pose, since we maintain the same relative transformation between the part and the gripper from pickup to assembly. 

Next, we determine the pickup position for all parts. Since all pickup motions follow a top-down path, parts are arranged to prevent any overlap along the Z-axis, the vertical direction in the world coordinate frame. This constraint ensures unobstructed pickup paths and simplifies the design of the supporting fixture. Along the Z-axis, parts are positioned at the lowest possible height while ensuring that there are no collisions with the ground based on their orientation. In addition, a minimum base height is maintained for the fixture board to provide structural stability and support. 

Determining the XY positions of the parts is more challenging, as the layout directly impacts the fixture area, which should ideally be minimized to reduce material costs for fixture fabrication and maximize the available workspace within the workcell. 
Additionally, incorrect part layout on the XY plane can lead to potential collisions between the gripper and the remaining parts during pickup. 
Since the orientation of each object is locked, we can represent each part using a rectangular bounding box of its 2D-projected contour.
Then, the problem becomes a 2D bin-packing problem, a classic problem with both heuristic and exact algorithms exist \cite{lodi2002two}.
We use a simple algorithm that iterates through the following phases: 
\begin{enumerate}
    \item We use the Maximal Rectangles algorithm \cite{jylanki2010thousand} to pack the bounding boxes into an initial bounding area; 
    \item We check the collisions between the gripper and the precedent parts of the grasped part; 
    \item If any collisions are detected, the colliding parts are buffered with additional spacing and the bin-packing process is performed again;
    \item We increase the bin area once it is not enough to find a packing solution given the increased rectangle sizes.
\end{enumerate}

Once an optimal packing configuration is determined, the fixture is generated by creating mold cavities that accommodate the part shapes. A minimal mold depth is calculated to ensure gravitational stability of part placement, where the Z-axis projection of each part's center of mass lies within the convex hull of the contact points between the fixture and the part. The fixture cavity is generated by projecting the part's 3D geometry onto a 2D plane perpendicular to the pickup direction and extruding it to the calculated depth. Additional cavity is generated by creating free space for the grasp motion for every part based on the gripper geometry and the grasping motion. Finally, the generated fixture is enhanced with countersunk pads to assist in accurate positioning. By automating the entire fixture generation process, our approach provides a flexible and scalable solution for diverse part geometries and assembly sequences.

\section{Experimental Setup}
\label{app:setup}

\subsection{Hardware Setup}

We conduct real-world experiments using a dual-arm setup composed of two Franka Emika Panda robots, each equipped with parallel-jaw grippers. The arms are mounted on one side of a shared table, facing the user, and their relative positions are calibrated via a common reference frame. The workspace is divided into a pickup area and an assembly area, both fixed and pre-defined based on the available workspace area. The system uses internal encoders for joint sensing, without external force-torque sensors or visual feedback.

\subsection{Simulation Environment}

We use RedMax, the same simulator used in \citet{tian2024asap}, for simulation-based planning, and Isaac Gym for reinforcement learning. Grasp feasibility is determined by sampling 100 candidate grasps per part using an antipodal grasp planner, followed by inverse kinematics validation and collision checking in RedMax. Precedence tier planning uses a physics-based disassembly planner~\cite{tian2022assemble} with orthogonal force directions for motion planning.

\subsection{Training Configuration}

Assembly policies are trained using PPO from the RL Games framework with key hyperparameters for RL presented in Table~\ref{tab:hyperparameters}, which we use for all reported RL experiments. Our generalist policies are trained for a maximum of $5\times10^7$ steps (or equivalently 1500 iterations) with a parallel rollout setup using 1024 environments, which takes less than 2 hours on a single NVIDIA RTX A6000 GPU. 

\begin{table}[h]
    \centering
    \caption{Key RL Hyperparameters.}
    \begin{tabular}{|l|l|}
        \hline
        \textbf{Parameter}               & \textbf{Value} \\ \hline
        Algorithm Name                   & PPO \\ \hline
        MLP Units                        & [256, 128, 64] \\ \hline
        MLP Activation                   & elu \\ \hline
        Learning Rate                    & 1e-4 \\ \hline
        Gamma                            & 0.99 \\ \hline
        Tau                              & 0.95 \\ \hline
        Entropy Coefficient              & 0.003 \\ \hline
        Gradient Norm                    & 1.0 \\ \hline
        Horizon Length                   & 32 \\ \hline
        Minibatch Size                   & 512 \\ \hline
        Mini Epochs                      & 8 \\ \hline
        Critic Coefficient               & 2 \\ \hline
        KL Threshold                     & 0.016 \\ \hline
    \end{tabular}
    \label{tab:hyperparameters}
\end{table}

\subsection{Real-World Deployment}

Robot control alternates between executing planned transit motions in joint space and reactive policy execution for insertion steps. For policy execution, a task-space impedance controller is used with gains $K_p =$ [800 N/m, 800 N/m, 400 N/m] and $K_d = 2\sqrt{K_p}$, transformed into the path-centric frame to ensure consistent compliance. Control runs at 30 Hz. During deployment, we use the leaky Policy-Level Action Integrator (leaky PLAI) scheme for improved stability with an action scale of 0.001 and an error threshold of 0.02. We use the same set of control parameters across all benchmark assemblies. Interventions are requested after three consecutive failures to insert, detected via joint deviation thresholds and motion stagnation.

\section{Ablation Studies}
\label{app:ablation}

\subsection{Impact of Coordinate and Action Design}

We conducted ablation studies to evaluate the impact of coordinate frame selection (world vs. path-centric) and action representation (nominal vs. residual) on the assembly specialist (AS) policy's performance in simulation. Table~\ref{tab:ablation_sim} presents the average \% of successful steps without intervention across 1024 randomized simulation evaluations for each assembly task under four configurations.

Overall, the combination of path-centric coordinates and residual actions proves particularly effective for assemblies involving diverse insertion directions, such as the Plumbers Block and Gamepad. Individually, each technique provides a structured prior that simplifies learning: path-centric coordinates standardize insertion directions by reorienting each assembly motion into a canonical frame, while residual actions leverages the planned trajectory and allow for fine-grained corrective adjustments on top of it. When applied together, they complement each other, enabling both directional consistency and precise control, leading to significantly higher success rates across most assemblies in our benchmark.

\begin{table*}[t]%
\small
\caption{Ablation studies on the impact of coordinate and action design choices on the assembly specialist (AS) policy's performance across 1024 randomized simulation evaluations.}
\label{tab:ablation_sim}
\begin{center}
\begin{adjustbox}{width={\textwidth}}
\begin{tabular}{cc|ccccccc}

\toprule
\multirow{2}{*}{\textbf{Coordinate}} & \multirow{2}{*}{\textbf{Action}} & \multicolumn{7}{c}{\textbf{\% of Successful Steps without Intervention (Simulation)}} \\ 
& & Beam & Plumber Block & Car & Gamepad & Cooling Manifold & Duct & Stool \\
\midrule
World & Nominal
& \textbf{99.71\%} & 49.32\% & \textbf{73.24\%} & 76.95\% & \textbf{94.92\%} & \textbf{96.97\%} & 70.41\% \\
World & Residual
& \textbf{99.71\%} & 95.02\% & 71.09\% & 73.44\% & \textbf{95.02\%} & \textbf{97.75\%} & 74.80\% \\
Path-Centric & Nominal
& \textbf{99.71\%} & 93.95\% & 60.35\% & 85.35\% & 92.58\% & \textbf{97.17\%} & 70.41\% \\
Path-Centric & Residual
& \textbf{99.12\%} & \textbf{97.46\%} & 70.12\% & \textbf{88.87\%} & \textbf{95.02\%} & \textbf{96.58\%} & \textbf{76.66\%} \\
\bottomrule

\end{tabular}
\end{adjustbox}
\end{center}
\end{table*}%

\subsection{Effect of Number of Trials}

In real-world experiments, we evaluate the assembly specialist policy with varying numbers of allowed trials (1, 2, or 3) per step, due to the stochastic nature of the policy. Table~\ref{tab:success_real} summarizes the average \% of successful steps without intervention across three end-to-end multi-step assembly runs. Results indicate that permitting additional trials substantially boosts success rates. The policy consistently improves as trial count increases, suggesting that the policy benefits from repeated attempts to refine alignment and correct minor positional errors.

\begin{table*}[t]%
\small
\caption{Scaling effect of number of policy trials across 3 complete end-to-end multi-step real-world evaluations.}
\label{tab:success_real}
\begin{center}
\begin{adjustbox}{width={\textwidth}}
\begin{tabular}{cc|ccccccc}

\toprule
\multirow{2}{*}{\textbf{Method}} & \multirow{2}{*}{\textbf{\# Trials}} & \multicolumn{7}{c}{\textbf{\% of Successful Steps without Intervention (Real World)}} \\ 
& & Beam & Plumber Block & Car & Gamepad & Cooling Manifold & Duct & Stool \\
\midrule
Open-Loop Tracking 
& 1 & 42\% & 25\%  & 20\% & 20\% & 17\% & 14\% & 21\% \\
\midrule
\multirow{3}{*}{\shortstack[c]{Assembly Specialist \\ Policy (AS)}}
& 1            & 58\% & 58\% & 60\% & 40\% & 67\% & 43\% & 75\% \\
& 2            & 67\% & 58\% & 80\% & 60\% & 67\% & 79\% & 75\% \\
& 3            & 75\% & 67\% & 87\% & 73\% & 83\% & 79\% & 88\% \\
\bottomrule

\end{tabular}
\end{adjustbox}
\end{center}
\end{table*}%

The results further demonstrate open-loop baseline's inability to recover from failures and adapt to variations in the real world environment, even though the planned path is accurate and the real robot is well calibrated. In contrast, the RL policy consistently demonstrates improved progress, with notable gains observed as the number of trials increases. The results show a clear trend: with 1 trial per step, the policy can achieve moderate progress but still faces challenges in most of the assemblies. Introducing retries significantly enhance progress, enabling the system to correct minor errors and overcome small disturbances.

\section{Integrating Vision Feedback: VLM for Insertion Alignment}

To further address insertion misalignments observed during the initial insertion attempt, we integrate a vision-language model (VLM) to provide corrective alignment feedback in the form of discrete actions. The model, a state-of-the-art version of Gemini (\texttt{gemini-2.5-pro-preview-03-25}), is leveraged to assess spatial alignment between the grasped part and the insertion hole based on visual input.

While training visuomotor policies directly from visual input is a common approach for alignment tasks, it requires extensive data collection, task-specific training, and continuous fine-tuning to generalize across diverse parts and insertion scenarios. In contrast, leveraging a VLM for alignment feedback offers significant advantages. VLMs are pre-trained on diverse visual contexts, enabling them to generalize across varied geometries and occlusions without extensive task-specific data collection. Additionally, they provide interpretable, discrete corrective actions (e.g., “move right”) accompanied by concise explanations, enhancing both robustness and transparency in alignment tasks, especially under low-cost, fixed-focus camera setups.

\subsection{Physical Setup}

\begin{wrapfigure}{r}{0.4\textwidth}
    \vspace{-12mm}
    \centering
    \includegraphics[width=\linewidth]{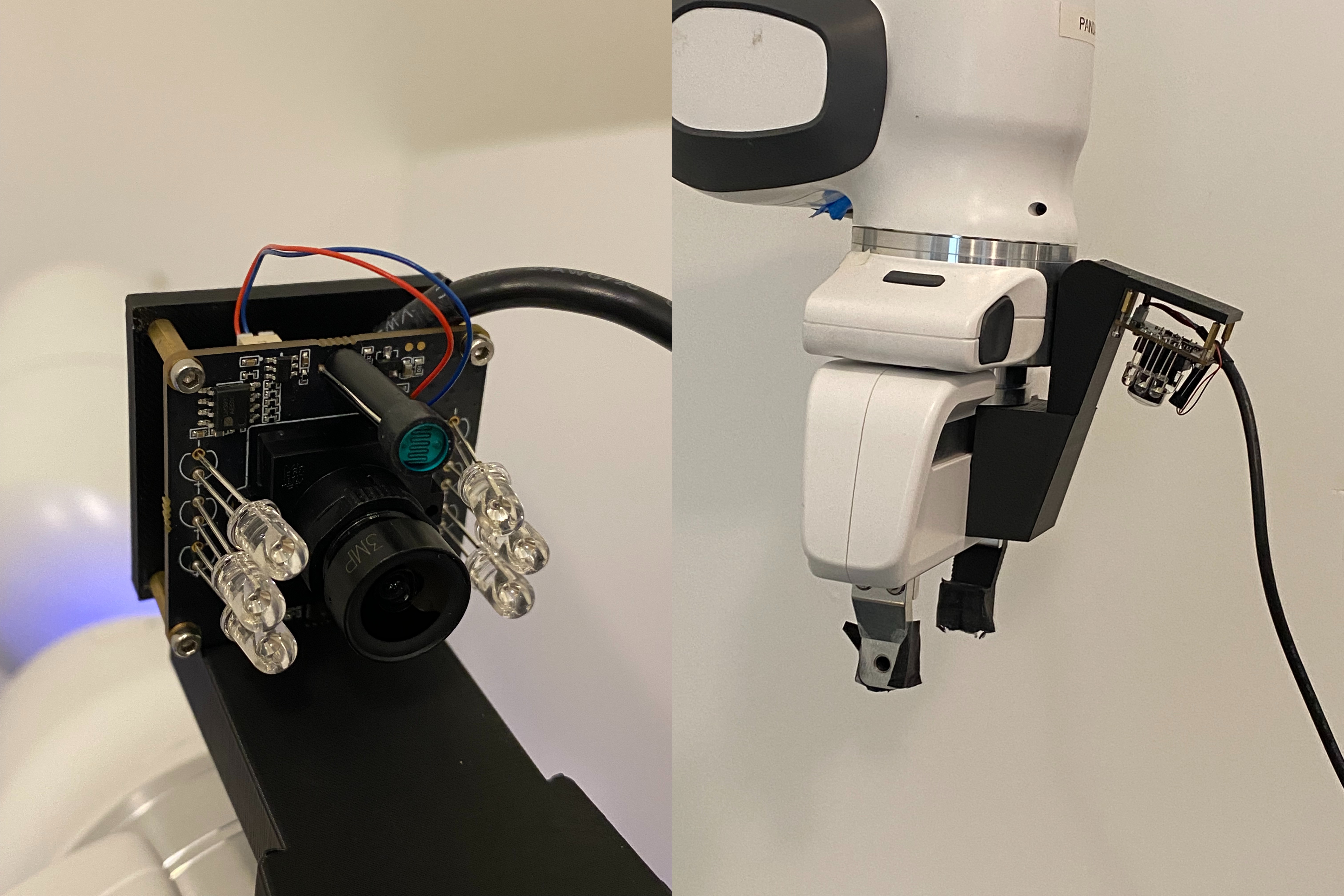}
    \caption{Physical setup for integerating vision feedback. Left: Camera details. Right: The mounted configuration on the robot wrist.}
    \label{fig:vision}
    \vspace{-8mm}
\end{wrapfigure}

The vision integration requires only RGB input without high imaging quality, allowing for the use of low-cost cameras. We utilize an Arducam B0205 USB camera (\$34.99), shown in Fig.~\ref{fig:vision}, mounted on the robot wrist at an angle optimized to capture the insertion area. The camera is equipped with an IR-CUT filter and infrared LEDs for low-light conditions, but lacks focus adjustment, resulting in reduced image clarity when the insertion part is out of focus.

\subsection{Methodology}
If the insertion policy fails on the first attempt, the entire video of the failed attempt is recorded and segmented into ten key frames sampled uniformly across the video. These frames are passed to the VLM, which is prompted to recommend the best corrective action (\texttt{up}, \texttt{down}, \texttt{left}, or \texttt{right}) to align the part with the hole. The VLM is additionally prompted to provide a concise, step-by-step reasoning for the recommended action to mitigate potential hallucinations. The exact prompt we use is detailed below.

\begin{lstlisting}[basicstyle=\ttfamily\small, breaklines=true]
'''
You are assisting a robot in aligning a grasped part for insertion using visual feedback from a camera mounted on the robot's wrist.


Task:
- The part is grasped by the robot and can move in four directions: ["up", "down", "left", "right"], each by 2 mm in the camera frame.
- The goal is to move the part to align it precisely with the hole for insertion.


Instructions:
- Carefully observe the video frames. Focus only on the position of the part relative to the hole.
- Determine the single best action to move the part to align with the hole.
- Focus only on spatial cues: Is the part too far left, right, above, or below the hole?


Response format:
{
"action": "right",
"reason": "The part is too far left relative to the hole and needs to move right to align."
}


Only output the single best action based on spatial cues. If the part is already aligned, output "hold".
What is the best action to move the part to align with the hole?
'''
\end{lstlisting}

After receiving the VLM feedback, the robot executes the recommended action by adjusting the gripper in the task frame by 2 mm. The insertion policy is then restarted from this new position. If the insertion still fails, the entire process is repeated with the newly recorded video sequence, allowing for multiple VLM interventions as necessary.

\subsection{Results and Analysis}

\begin{figure*}[h]
    \centering
    \includegraphics[width=\textwidth]{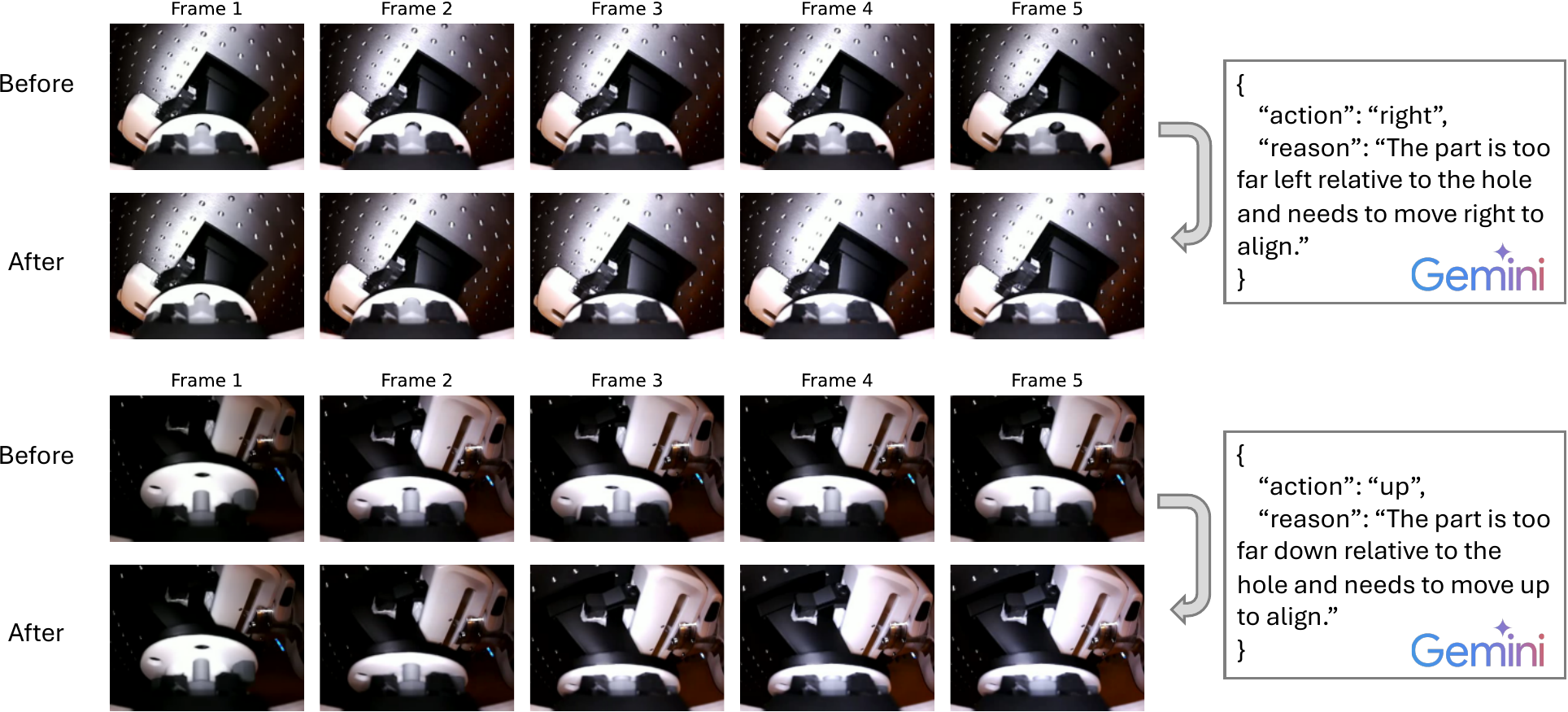}
    \caption{Example outputs from VLM during corrective alignment. The VLM identifies spatial misalignments in the camera frame and recommends discrete corrective actions (e.g., “right” and “up”) with concise reasoning.}
    \label{fig:gemini}
\end{figure*}

The VLM integration demonstrated notable improvements in alignment accuracy, particularly in cases where the insertion policy initially failed due to misalignment. Figure~\ref{fig:gemini} illustrates two representative examples where the VLM provided corrective actions that successfully guided the arm to the intended alignment.

In the first example, the VLM identified a leftward misalignment and recommended the action “right”, allowing the part to be re-centered relative to the insertion hole. The corrective action was executed in the tool frame, resulting in a more precise alignment before the subsequent insertion attempt. Similarly, in the second example, the VLM detected a downward offset and suggested the action “up”, effectively repositioning the part closer to the target insertion point. In both cases, a single VLM intervention was sufficient to resolve the misalignment, highlighting the model’s capacity to reason spatially based on minimal visual input.

Despite the low-cost camera setup and lack of focus adjustment capabilities, the VLM effectively discerned alignment cues based on coarse visual features. This is particularly noteworthy given that occlusions and visual clutter are prevalent in multi-part assemblies, where small positional errors can accumulate over successive steps. The VLM’s concise, reason-based output structure further mitigates hallucination risks by constraining the response format to a single action-reason pair, reducing ambiguity and enhancing interpretability.

\subsection{Limitations and Future Work}
While the VLM integration demonstrated significant alignment improvements, occlusions remained a primary failure mode. Occlusions are common in complex assemblies, necessitating either a flexible/active camera setup or multiple cameras strategically positioned to cover the scene comprehensively. Furthermore, hallucination remains a concern, particularly in cluttered scenes where visual cues are ambiguous. Future work will explore improving prompt engineering and camera configurations, potentially leveraging multi-view setups and active camera movements akin to human head and body movements.

\end{appendices}

\end{document}